\pdfoutput=1
\documentclass[11pt]{article}

\usepackage{ACL2023}
\usepackage{times}
\usepackage{latexsym}

\usepackage[T1]{fontenc}
\usepackage[utf8]{inputenc}

\usepackage{microtype}

\usepackage{inconsolata}
\usepackage{microtype}
\usepackage{color}
\usepackage{booktabs}
\usepackage{multirow}
\usepackage{amsmath}
\usepackage{graphicx}
\usepackage{float} 
\usepackage{makecell}
\usepackage{CJKutf8}
\usepackage{xcolor}
\usepackage{soul}
\usepackage{color}
\usepackage{comment}
\usepackage{mathrsfs}
\usepackage{threeparttable}
\usepackage{bm}
\usepackage{amsfonts}

\title{StoryTrans: Non-Parallel Story Author-Style Transfer with Discourse Representations and Content Enhancing}

\author{
 Xuekai Zhu$^2$\thanks{\ \ Equal contribution. }, Jian Guan$^1$\footnotemark[1] , \textbf{Minlie Huang}$^1$\Thanks{~Corresponding author} and Juan Liu$^2$\\
\small $^1$The CoAI group, DCST, Institute for Artificial Intelligence, State Key Lab of Intelligent Technology and Systems, \\
 \small $^1$Beijing National Research Center for Information Science and Technology, Tsinghua University, Beijing 100084, China \\
\small $^2$ Institute of Artificial Intelligence, School of Computer Science, Wuhan University, Wuhan, 430072, China\\
\small \texttt{\{xuekaizhu,liujuan\}@whu.edu.cn},
\small{\texttt{j-guan19@mails.tsinghua.edu.cn}}, 
\small{\texttt{aihuang@tsinghua.edu.cn}}
 \\
}

\begin{document}
\maketitle
\begin{abstract}
Non-parallel text style transfer is an important task in natural language generation. However, previous studies concentrate on the token or sentence level, such as sentence sentiment and formality transfer, but neglect long style transfer at the discourse level. Long texts usually involve more complicated author linguistic preferences such as discourse structures than sentences. In this paper, we formulate the task of non-parallel story author-style transfer, which requires transferring an input story into a specified author style while maintaining source semantics. To tackle this problem, we propose a generation model, named StoryTrans, which leverages discourse representations to capture source content information and transfer them to target styles with learnable style embeddings. We use an additional training objective to disentangle stylistic features from the learned discourse representation to prevent the model from degenerating to an auto-encoder. Moreover, to enhance content preservation, we design a mask-and-fill framework to explicitly fuse style-specific keywords of source texts into generation. Furthermore, we constructed new datasets for this task in Chinese and English, respectively. Extensive experiments show that our model outperforms strong baselines in overall performance of style transfer and content preservation.
\end{abstract}

\section{Introduction} \label{1}
Text style transfer aims to endow a text with a different style while keeping its main semantic content unaltered.
% ~\cite{hovy1987generating}.
% Text style transfer aims to convert the stylistic properties of a text while preserving its main semantics
It has a wide range of applications, such as formality transfer \cite{jain2019unsupervised}, sentiment transfer \citep{10.5555/3295222.3295427} and author-style imitation \citep{tikhonov2018guess}.

\begin{figure}[ht] 
\centering 
\includegraphics[width=0.98\linewidth]{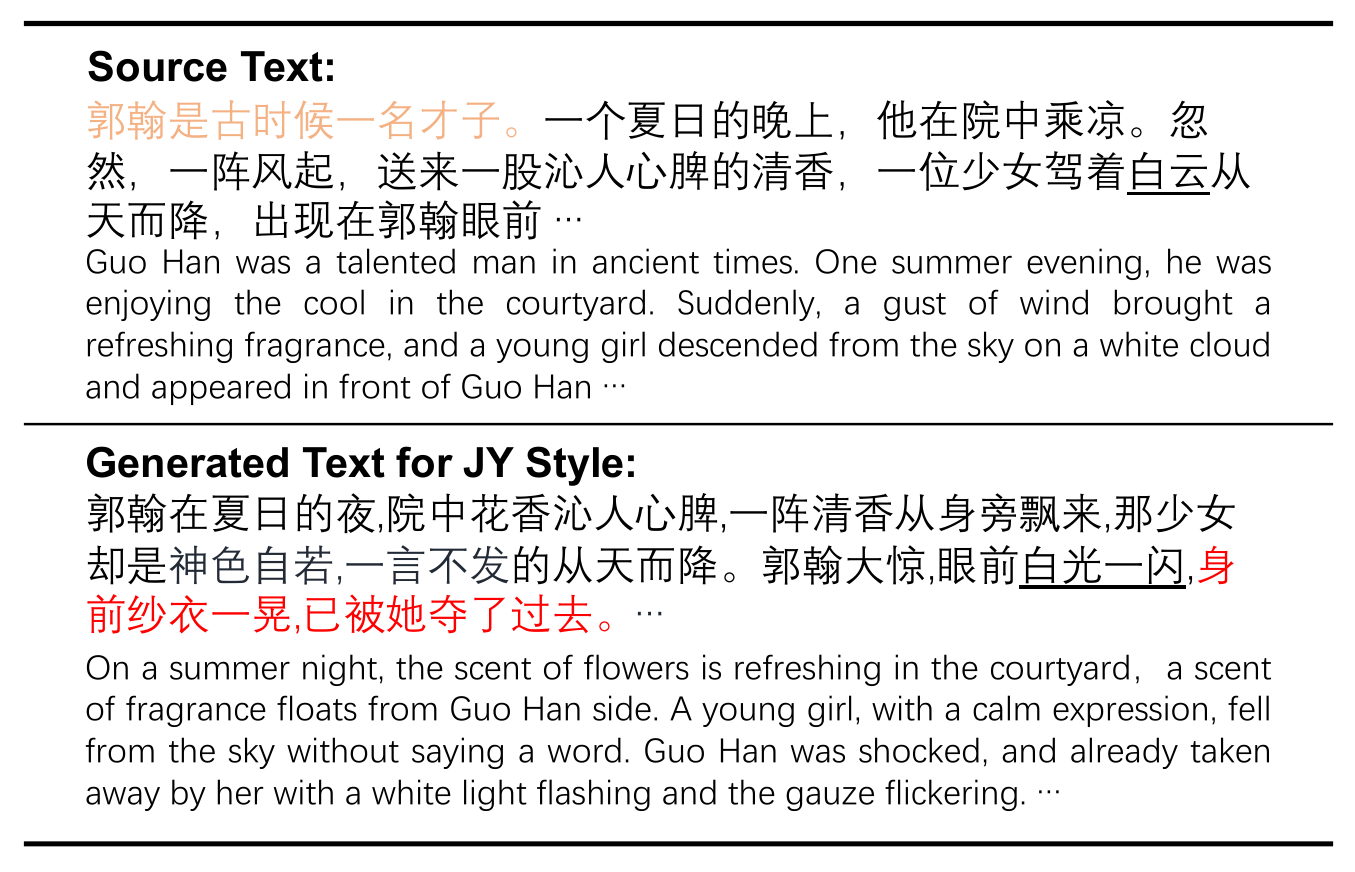} 
\captionsetup{type=table}
\caption{An example that transfers a vernacular story
% in the LOT dataset~\cite{guan2021lot} 
to the martial arts style of JY generated by StyleLM. The orange sentence indicates missing content in source text.
% The \textbf{bold} sentences correspond to rewritten contents. 
The rewritten token is underlined.
The red highlights are supplementary short phrases or plots to align with the target style. The English texts below the Chinese are translated versions of the Chinese samples.}
\label{transfered_sample} 
\end{figure}

Due to the lack of parallel corpora, recent works mainly focus on unsupervised transfer by self-reconstruction.
% fusing the target style signals and the representations of source contents. 
%Concretely, 
Current methods proposed to disentangle styles from contents by removing stylistic tokens from inputs explicitly~\cite{huang-etal-2021-nast} or reducing stylistic features from token-level hidden representations of inputs implicitly~\cite{lee-etal-2021-enhancing}. This line of work has impressive performance on single-sentence sentiment and formality transfer. However, it is yet not investigated to transfer author styles of long texts such as stories, manifesting in the author's linguistic choices at the lexical, syntactic, and discourse levels. 
% Specifically, the critical issue of mainstream work \citep{ijcai2019-732, john-etal-2019-disentangled, dai-etal-2019-style, he2020a, lee-etal-2021-enhancing, huang-etal-2021-nast} is hard to reveal the insight of narrative techniques or syntactic elements transfer, which lies in sentence-level or even document-level linguistic features.

In this paper, we present the first study on 
story author-style transfer, which 
aims to rewrite a story incorporating source content and the target author style. {\textbf{The first challenge} of this task lies in imitation of author's linguistic choices at the discourse level, such as {narrative techniques~(e.g., brief or detailed writing)}}.
% and syntactic elements}}. %, condensed or expanded content}) beyond token-level transfer. 
As exemplified in Table~\ref{transfered_sample}, the generation text for the JinYong~(JY)\footnote{JinYong is a Chinese martial arts
novelist.}
%and essayist.} 
style not only rewrites %source sentences to the martial arts style~(e.g., sentences 2 and 3), 
some tokens to the martial arts style~
(e.g., \begin{CJK*}{UTF8}{gkai}{“白云”}\end{CJK*}~/“white cloud” to \begin{CJK*}{UTF8}{gkai}{“白光一闪”}\end{CJK*}~/“light flashing”)
% (e.g., ``clam expression'', ``gauze flickering'' in generated text behind sentence 2,3),
but also adds additional events in detail and enrich the storyline~(e.g., the red highlights).
% behind sentence 4 and 5).
% corresponding to target narrative style.
% describe character plot in more detailly, 
% forming a coherent event sequence of 
% the target narrative style
%weave many details of the martial arts style into original plots of the source text.
%prefers to portray the plot of the story, and is rephrased into a variety of martial arts action descriptions (in red highlights). %Comparing the generation text with the source text in Table \ref{transfered_sample}, 
%For example, content of sentences 2 and 3 are merged while syntactic elements are rephrased, sentence 4 is rewritten into colloquial expression and lost some content, and sentence 5 is replaced the subject and object. 
%These sentences are rewritten in terms of narrative techniques or syntactic elements and meanwhile keep consistent in main semantics with source texts. 
In contrast to the transfer of token-level features like formality, it is more difficult to capture the inter-sentence relations correlated with author styles and disentangle them from contents. %in terms of inter-sentence relations are harder to capture and disentangle from contents.
\textbf{The second challenge} is %the high specificity between author styles and contents, which means different author styles tend to be associated with specific writing topics. 
that the author styles tend to be highly associated with specific writing topics.
Therefore, it is hard to transfer these style-specific contents to another style. For example, the topic ``talented man'' hardly shows up in the novels of JY, leading to the low content preservation of such contents, as shown in the orange text in Table~\ref{transfered_sample}.
%topic difference between texts of different authors~(i.e., \textit{topic drift}), which makes it hard to preserve the content information of source texts. 
%As exemplified in the Table \ref{transfered_sample} green sentence, it is difficult for a language model trained on the corpus of martial art novels to combine his style and the topic "talented man" that the model has never seen, while it is high frequency in idiom story corpus \cite{guan2021lot}. 

To alleviate the above issues, we propose a generation framework, named \textbf{StoryTrans}, which learns discourse representations from source texts and then combines these representations with learnable style embeddings to generate texts of target styles. Furthermore, we propose a new training objective to reduce stylistic features from the discourse representations, which aims to pull the representations derived from different texts close in the latent space. To enhance content preservation, we separate the generation process into two stages, which first transfers the source text with the style-specific content keywords masked and then generates the whole text by imposing these keywords explicitly.

To support the evaluation of the proposed task, we collect new datasets in Chinese and English based on existing story corpora.\footnote{The codes and data are available at \url{https://github.com/Xuekai-Zhu/storytrans_public}}
We conduct extensive experiments to transfer fairy tales~(in Chinese) or everyday stories~(in English) to typical author styles, respectively. 
Automatic evaluation results show that our model achieves a better overall performance in style control and content preservation than strong baselines. The manual evaluation also confirms the efficacy of our model.
We summarize the key contributions of this work as follows:

\noindent\textbf{I.} To the best of our knowledge, we present the first study on story author style transfer. We
    construct new Chinese and English datasets for this task.
    
    \noindent\textbf{II.} We propose a new generation model named StoryTrans to tackle the new task, which implements content-style disentanglement and stylization based on discourse representations, then enhances content preservation by explicitly incorporating style-specific keywords.
      % which learns 
    % % style-independent 
    % discourse representations to capture the content information and 
    
    \noindent\textbf{III.} Extensive experiments show that our model outperforms baselines 
    in the overall performance of style transfer accuracy and content preservation.
%\end{itemize}

\section{Related Work}

\begin{figure*}[ht] 
\centering 
\includegraphics[width=0.9\linewidth]{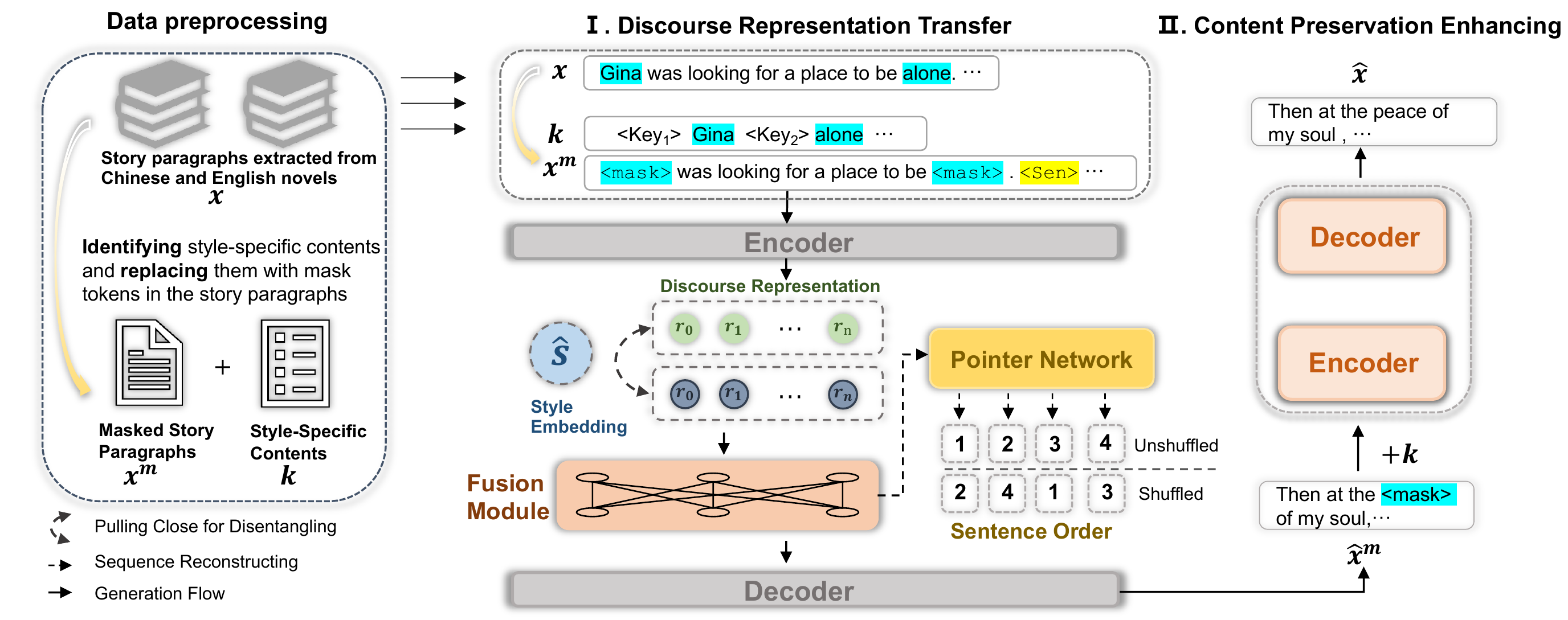} 
\caption{An overview of the generative flow. For discourse representation transfer (the first stage), the encoder employs discourse representations ($\{\bm{r}_i\}_{i=1}^n$) to contain main semantics of pre-processed input ($\bm{x^m}$). Then, the fusion module stylizes the discourse representations with target style embedding (${\hat{s}}$). For content preservation enhancing (the second stage), our model enhances the content preservation of transferred texts ($\bm{x^m}$) with style-specific content ($k$). $\bm{x}$ and $\bm{\hat{x}}$ denote the original story and the final output, respectively.}
\label{Fig_1} 
\end{figure*}

\subsection{Style Transfer}
Recent studies concentrated mainly on token-level style transfer of single sentences, such as formality or sentiment transfer. 
We categorize these studies into three following paradigms. 

The first paradigm built a style transfer system without explicit disentanglement of style and content. This line of work used additional style signals or a multi-generator structure to control the style. 
% on an attention-based encoder-decoder model and controls the style by style embedding or multi-generator structures  
\citet{dai-etal-2019-style} added an extra style embedding in input for manipulating the style of texts. \citet{ijcai2020-526} proposed a style instance encoding method for learning more discriminative and expressive style embeddings. The learnable style embedding is a ﬂexible yet effective approach to providing style signals. Such a design helps better preserve source content. \citet{syed2020adapting} randomly dropped the input words, then reconstructed input for each author separately, which obtained multiple author-specific generators. The multi-generator structure is effective but also resource-consuming. 
% since each style is represented as the generator parameters. 
% In contrast to multi-generator structures, 
However, this paradigm incurs unsatisfactory style transfer accuracy without explicit disentanglement.

The second paradigm 
% disentangled the style from content explicitly. This 
% paradigm 
disentangled the content and style explicitly in latent space, then combined the target style signal.
%with the style-independent content representation.
\citet{zhu-etal-2021-neural} 
% disentangled the content and style by 
diluted sentence-level information in style representations. \citet{john-etal-2019-disentangled} incorporated style prediction and adversarial objectives for disentangling.
%the latent representations. 
%  \citet{gao-etal-2019-structuring}
% devised a set of training objectives to achieve the disentanglement. 
% 
\citet{lee-etal-2021-enhancing} removed style information of each token with reverse attention score~\cite{bahdanau2015neural}
, which is estimated by a pre-trained style classifier.
% employed knowledge attained~\cite{bahdanau2015neural} to suppress original style signal at token representation, and conditional layer normalization to add target style signal. 
This paradigm utilizes adversarial loss functions or a pre-trained estimator for disentanglement. And experiment results indicate that explicit disentanglement leads to satisfactory style transfer accuracy but poor content preservation.
% control ability. 
% Our LongTrans extends above ideas of disentanglement to discourse representations.
% These works illustrate style and content have improper entanglement in latent space. Explicit modeling can lend models stable style control ability.

The final paradigm views style as localized features of tokens in a sentence, which locates style-dependent words and replaces the target-style ones.  \citet{xu-etal-2018-unpaired} employed an attention mechanism to identify style tokens and ﬁlter out such tokens. \citet{ijcai2019-732} utilized a two-stage framework to mask all sentimental tokens and then infill them. \citet{huang-etal-2021-nast} aligned words of input and reference to achieve token-level transfer. To sum up, this paradigm maintains all word-level information, but it is hard to apply to the scenarios where styles are expressed beyond token level, e.g., author style.

Absorbing ideas from paradigm 1 and 2, we apply explicit disentanglement by pulling close discourse representations, which is formulated into disentanglement loss. Furthermore, we design a fusion module to stylize the 
% style-independent 
discourse representation. 

\subsection{High-Level Representation}
% Impressive advances have been witnessed in many NLP tasks with pretrained contextualized representation. However
% most models were limited on token-level representation learning, which is not enough for capturing the hierarchical structure of natural language texts.
Prior works captured the hierarchical structure of natural language texts by learning high-level representations.
% generate coherent text by modeling high-level representation, which aims 
% \citet{li-etal-2015-hierarchical} built an hierarchical embedding for a paragraph, and learned a RNN-based decoder by reconstructing the original paragraph. 
% % \citet{kiros2015skip} tried to reconstruct neighboring sentences while encoding a sentence.
% \citet{zhang-etal-2019-hibert} proposed to learn sentence representations by recovering masked sentences.
\citet{li-etal-2015-hierarchical} and \citet{zhang-etal-2019-hibert} proposed to learn hierarchical embedding representations by reconstructing masked version of sentences or paragraphs.
% based on a hierarchical architecture.
\citet{reimers-2019-sentence-bert} derived semantical sentence embeddings by fine-tuning BERT~\cite{devlin-etal-2019-bert} on downstream tasks. 
\citet{lee-etal-2020-slm,guan-etal-2021-long} inserted special tokens for each sentence and devised several pre-training tasks to learn sentence-level representations. 
% However, to the best of our knowledge, there is no prior research on long text style transfer. 
We are inspired to use a sentence order prediction task to learn high-level discourse representations.
%, which is formulated into a sequence order loss. }
% And more related works are presented in Appendix~\ref{more_related_work}

% \section{More Related Work}\label{more_related_work}
\subsection{Long Text Generation} \label{2.3}
In order to generate coherent long texts,
recent studies usually decomposed generation into multiple stages. \citet{fan-etal-2018-hierarchical, yao2019plan} generated a premise, then transformed it into a passage.
% \citet{yao2019plan} planned a storyline, and then generated a story based on the storyline. 
\citet{tan-etal-2021-progressive} first produced domain-specific content keywords and then progressively refines them into complete passages. 
% This line of works first generated a rough sketch, and then extended it into the complete text with details. 
Borrowing these ideas 
% from above works
, we adopted a mask-and-fill framework to enhance content preservation in text style transfer.

\section{Methodology} \label{3}

\subsection{Task Definition and Model Overview}
We formulate the story author-style transfer task as follows: assuming that $S$ is the set of all author-styles, given a multi-sentence input $ \bm{x} = (x_1, x_2, \cdots, x_T)$ of $T$ tokens and its author-style label ${s}\in S$, the model should generate a multi-sentence text %$\bm{y} = (y_1, y_2, \cdots, y_j)$ of $j$ tokens 
with a specified author-style ${\hat{s}}\in S$ while keeping the main semantics of $\bm{x}$.

%As mentioned in section 1, 
%To tackle the style transfer problem we deﬁned above, we propose xxxxxx
%divide the LTAST into two sub-tasks, Srt and Cpe.
%Specifically, Our model aims to transfer author-style of input, and Cpe tries to enhance content preservation, which means Srt and Cpe conduct on different training goals. Note that Srt and Cpe are trained separately, but the full generative flow is comprised of Srt and Cpe. 
%However, 
%As aforementioned, %to improve content preservation, 

As illustrated in Figure~\ref{Fig_1}, we split the generation process into two stages. %Concretely, we replace $\bm{k}$ with special mask token, <mask>. The details of identifying the important content would described in later subsection (\S\ref{3.3}).
% In data preprocessing, 
We first identify style-specific keywords $\bm{k}=(k_1,k_2,\cdots,k_l)$ from $\bm{x}$, and then mask them with special tokens $\langle\texttt{mask}\rangle$. We denote the resulting masked version of $\bm{x}$ as $ \bm{x^{m}} = (x_1^m, x_2^m, \cdots, x_T^m)$. 
In the first generation stage, we perform discourse representation transfer on $\bm{x^{m}}$. In the second stage, we complete the masked tokens in the output of the first stage conditioned on $\bm{k}$ in a style-unrelated manner. %to achieve enhancement of content preservation. 

\begin{figure}[t!]
\centering
\includegraphics[width=0.9\linewidth]{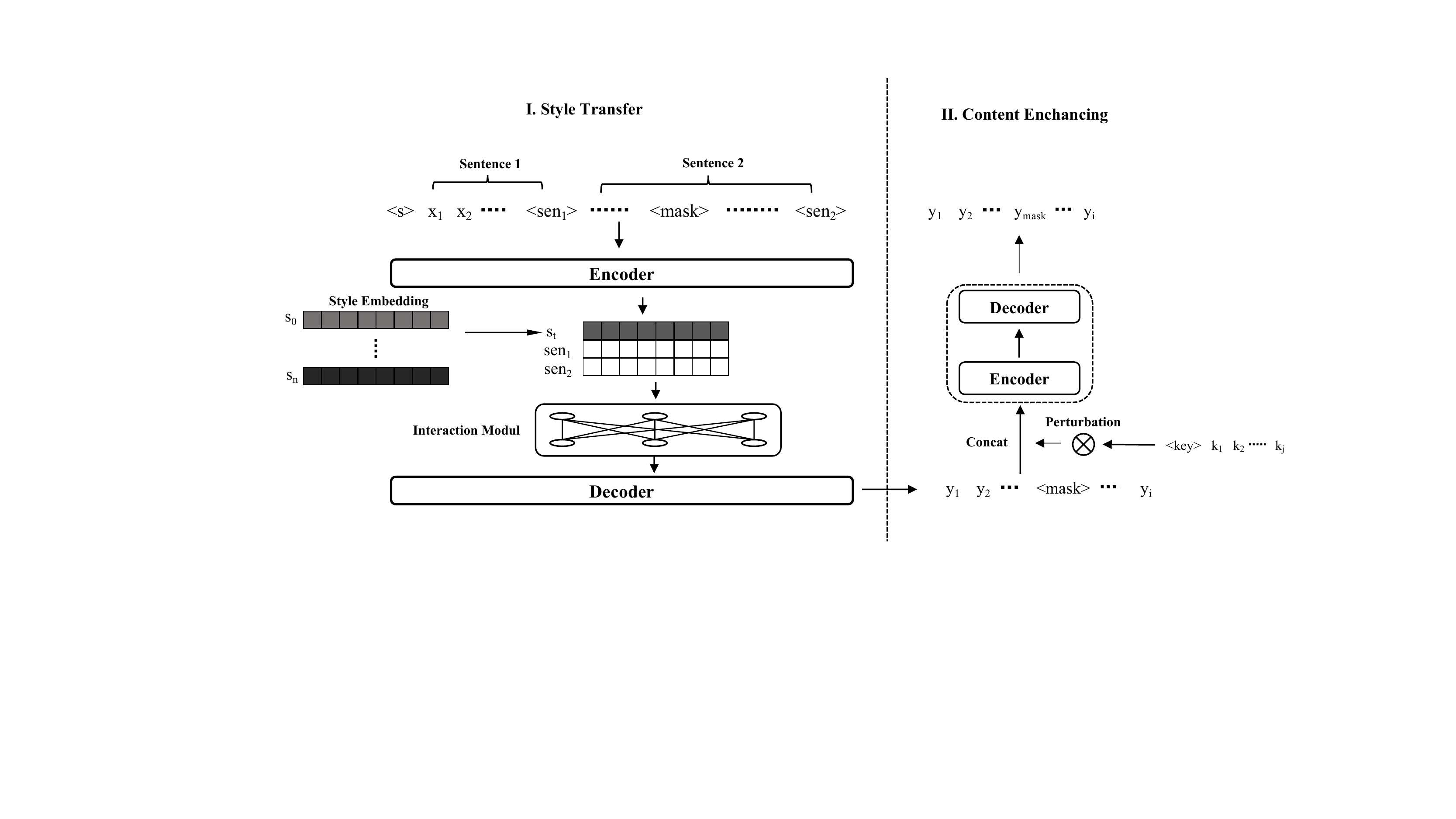}
\caption{Illustration of loss functions during training for the first stage~(a) and second stage~(b). %The a/b indicates first/second training stage. 
Enc, Fus, Dec and C denote the encoder, the fusion module, the decoder, and style classifier, respectively. 
% Note that encoder and decoder in (a) and (b) are different models.
} 
\label{training}
\end{figure}

Due to the lack of parallel data, typical style transfer models tend to optimize the self-reconstruction loss with the same inputs and outputs~\cite{xiao2021transductive,lee-etal-2021-enhancing}.
Obviously, training with only the self-reconstruction loss %standard language modeling loss 
will make the model easily ignore the target style signals and simply repeat the source inputs.
%Previous studies demonstrate that unsupervised models tend to back in vanilla auto-encoder without any regularization, which has no capability to control the style. 
Therefore, in the first stage, we devise an additional training objective, to disentangle stylistic features from intermediate discourse representations %, which obtain style-independent sentence representation 
$\{\bm{r}_i\}_{i=1}^n$, where $n$ is the number of sentences. Then, we fused these style-independent discourse representations with the target style $\hat{s}$ as a discourse-level guidance for the subsequent generation of the transferred text. As for discourse representations learning, we employ a sentence order prediction loss to capture inter-sentence discourse dependencies.
And we use a style classifier loss to control the style of generated texts~\cite{lee-etal-2021-enhancing}.
In summary, the fist-stage model is trained using the following loss function: 
\begin{align}
\mathcal{L}_1
% _{\rm stage_1} 
= \mathcal{L}_{\rm self} + \lambda_1\mathcal{L}_{\rm dis} + \lambda_2\mathcal{L}_{\rm sop}+\lambda_3\mathcal{L}_{\rm style}, \label{stage1}
\end{align} 
where $\lambda_1$, $\lambda_2$~and~$\lambda_3$ are adjustable hyper-parameters. $\mathcal{L}_{\rm self}$, $\mathcal{L}_{\rm dis}$, $\mathcal{L}_{\rm sop}$ and $\mathcal{L}_{\rm style}$ are the self-reconstruction loss, the disentanglement loss, the sequence order prediction loss and the style classifier loss, respectively. Figure~\ref{training} shows the workflow of learning objects. 
%To enable the style control, we insert an interaction module between the encoder and decoder, which aims to stylize the style-independent sentence representations with the target style embedding. Similar to the conventional generation models, encoder and interaction module are bidirectional networks, but decoder is a left-to-right generator. 
% $\mathcal{L}_{\rm gen}$: %estimates the conditional probability for the output sentence $\bm{y^m} = (y_1^m, y_2^m, \cdots, y_j^m)$ by auto-regressively factorized its as:
%  We train the decoder by minimizing the self-reconstruction loss:

In the second stage, we use a denoising auto-encoder~(DAE) loss to train another encoder-decoder model for reconstructing $\bm{x}$: %conditioned on $\bm{x^m}$ and $\bm{k}$: %to generate the final output, which also could be formulated as follows:
\begin{gather}
\mathcal{L}_2 = -\sum_{t=1}^T\text{log} P(x_t|x_{<t}, \{k_i\}_{i=1}^l, \bm{x}^{m}). \label{dae_loss}
%p(\bm{y}\mid \bm{y^m}, \bm{k})= \prod_{t=1}^j p(y_t\mid y_1,\cdots, y_{t-1}, \bm{z})
\end{gather} 
%where $\bm{z}$ is the contextualized representation of $\bm{y^m}$ and $\bm{k}$ acquired from the encoder.
This stage is unrelated to author styles, and helps achieve better content preservation. 

\subsection{Discourse Representations Transfer}
% \paragraph{Discourse Representations Transfer}
% With scarce parallel corpora, we conduct the long text author-style transfer in an unsupervised the environment, in which the ground truth is not provided. 
% Following previous unsupervised methods, our model aims to reconstruct the input for training. 

% In analogy to long text generation,
As described in Figure~\ref{training}, 
we propose to learn discourse representations,
and then reconstruct the texts from discourse representations. 
And we perform the disentanglement and stylizing operation based on discourse representations.
% condense semantics of multiple sentences to intermediate representations (sentence representations) 
% Since complicated stylistic features, the disentanglement and stylizing operation are performed on high-level representations.

\paragraph{Discourse Representations}
 Supposing that $\bm{x^m}$ consists of $n$ sentences,  
we insert a special token $\langle\texttt{Sen}\rangle$ at the end of each sentence in $\bm{x^m}$~\cite{reimers-2019-sentence-bert,lee-etal-2020-slm,guan-etal-2021-long}. 
Let $\bm{r}_n$ denote the hidden state of the encoder at the position of the $n$-th special token, $\{\bm{r}_i\}_{i=1}^n= {\rm Encoder}(\bm{x}^m)$.
And $\bm{z}_n$ is the output of the fusion module corresponding to $\bm{r}_n$. Previous studies have demonstrated that correcting the order of shuffled sentences is a simple but effective way to learn meaningful discourse representations~\cite{lee-etal-2020-slm}. %, which can be fformulated into sequence order loss. 
As shown in Figure~\ref{Fig_1}, we feed  $\bm{z}_n$ into a pointer network~\cite{gong2016end} to predict orders. 
% , which consists of one self-attention layer
%Then pointer network tries to predict the correct order from shuffled/unshuffled sentences.
% , which is formulated into order prediction loss. 
During training, we shuffled the original sentence order and feed the perturbed text into the encoder for calculating $\mathcal{L}_{\rm sop}$.

\paragraph{Fusion Module}
To provide signals of transfer direction, we concatenate the learned discourse representations $\{\bm{r}_i\}_{i=1}^n$ with the style embedding ${\bm{s}}$
% ~\cite{dai-etal-2019-style} 
and fuse them using a multi-head attention layer, %which is a simple and practical method from previous studies , 
%which can learn the style of text implicitly. 
as illustrated in Figure \ref{Fig_1}. To capture discourse-level features of texts with different author-styles, we set each style embedding to a vector with the same dimension as $\bm{r}_i$.
% a sequence of $n$ vectors. 
Formally, we derive the style-aware discourse representations $\{\bm{z}_i\}_{i=1}^{n+1}$ as follows:
%, we feed 
%the target style embedding $\bm{\hat{s}}$ and $\{sen_i\}_i^n$ are concatenated and then fed into interaction module. 
%Interaction module aims to get new sentence-level representations entangling target author style and source text semantics. The process of the interaction module is formulated as below:
\begin{gather}
%\{z_i^{\hat{s}}\}_i^{n+1} = {\rm Int}(\bm{\hat{s}}\parallel \{sen_i\}_i^n)
% \{\bm{z}_i\}_{i=1}^n=\text{MHA}(Q=K=V=\{\bm{r}_i\parallel{\bm{s}}_i\}_{i=1}^n),
\{\bm{z}_i\}_{i=1}^{n+1}=\text{MHA}(Q=K=V=\{{\bm{s}}\parallel\{\bm{r}_i\}_{i=1}^n\}),
\end{gather}
where MHA is the multi-head attention layer, Q/K/V is the corresponding query/key/value, $\parallel$ is the concatenation operation.
% , and ${\bm{s}}_i$ is the $i$-th vector in ${\bm{s}}$. %, $\{z_i\}_i^{n+1}$ is the stylized sentence representation acquired from the interaction module final hidden state. 
Then, the decoder gets access to $\{\bm{z}_i\}_{i=1}^{n+1}$ through the cross-attention layer, %we point the cross-attention layer in the decoder to $\{\bm{z}_i\}_{i=1}^n$, 
which serve as a discourse-level guidance for generating the transferred texts.
Then, we feed $\{\bm{z}_i\}_{i=1}^{n+1}$ into the decoder. 

\paragraph{Pointer Network} 
Following \citet{logeswaran2018sentence,lee-etal-2020-slm},
we use a pointer network to predict the original orders of the shuffled sentences. The each position probability of sentence order is formulated as follows:
\begin{gather}
    p_i = \text{softmax}(\{z_i\}_{i=1}^n W z_i^T),
\end{gather}
where $p_i$ is predicted probabilities of sentence $i$, $W$ is a trainable parameter. 

\subsection{First-Stage Training Objectives}
\paragraph{Self-Reconstruction Loss}
We formulate self-reconstruction loss as follows:
% The  can be formulated:
\begin{gather}
\mathcal{L}_{\rm self} = -\sum_{t=1}^T\text{log} P(x^m_t|x^m_{<t}, \{\bm{r}_i\}_{i=1}^n, {\bm{s}}),\label{self1}
%p(\bm{y^m}\mid \bm{x^m}, \bm{\hat{s}})= \prod_{t=1}^j p(y_t^m\mid y_1^m,\cdots, y_{t-1}^m, \{sen_i\}_i^n) 
\end{gather}
where ${\bm{s}}$ is the learnable embedding of ${s}$. During inference, we replace ${\bm{s}}$ with the embedding of the target style $\hat{{s}}$~(i.e., $\hat{\bm{s}}$), to achieve the style transfer.

\paragraph{Disentanglement Loss}
We disentangle the style and content on discourse representations.
Inspired by prior studies on structuring latent spaces~\cite{gao-etal-2019-structuring,zhu-etal-2021-neural}, we devise an additional loss function $\mathcal{L}_{\rm dis}$ %on the sentence representations 
to %regularize $\{\bm{r}_i\}_{i=1}^n$ 
pull close discourse representations from different examples in the same mini-batch, corresponding to different author styles. 
%in the content latent space between examples with different author-styles in the same mini-batch.
%Previous studies~\citep{gao-etal-2019-structuring,zhu-etal-2021-neural} propose to transfer style of response in dialog through regularizing latent representation around target style text in the structure latent space. Put simply, in order to generate stylized response, they try to align representation of target-style text and un-transferred text in latent space. To strip style information from sentence-level representation, we proposed sen loss $\mathcal{L}_{sen}$ to align sentence-level representations in the same training batch. 
%In summary, 
$\mathcal{L}_{\rm dis}$ and $\mathcal{L}_{\rm self}$ work as adversarial losses and lead the model to achieve a balance between content preservation and style transfer. %Specifically,
We derive $\mathcal{L}_{\rm dis}$ as follows:
\begin{align}
% \mathcal{L}_{\rm sen} &= \mathbb{E}_{\{S_i\}_i^k} \parallel S_{k_1} - S_{k_2} \parallel_2^2,\\
\mathcal{L}_{\rm dis} &= \frac{1}{2b} \sum_{i=1}^{b}\sum_{j=1}^{b} \parallel \Bar{\bm{r}}_i - \Bar{\bm{r}}_j \parallel_2^2, \\
\Bar{\bm{r}} &= \frac1n{\sum_{i=1}^n \bm{r}_i}
\end{align}
where $b$ is the size of mini-batch.

\paragraph{Sentence Order Prediction Loss}
We formulate $\mathcal{L}_{\rm sop}$ as the cross-entropy loss between the golden and predicted orders %of shuffled sentences and original ones, 
as follows:
\begin{align}
    \mathcal{L}_{\rm sop} = -\frac{1}{n}\sum_{i=1}^{n}o_i\log(p_i),
\end{align}
where $o_i$ is a one-hot ground-truth vector of correct sentence position, and $p_i$ is predicted probabilities.
% and $w_i$ is the output of the pointer network. 

\paragraph{Style Classifier Loss} We expect the transferred text to be of the target style. Hence we train a style classifier to derive the style transfer loss as follows:
\begin{align}
    \mathcal{L}_{\rm style} = -\mathbb{E}_{\hat{\bm{x}}^m\sim {\rm Decoder}}[\text{log}P_C(s|\hat{\bm{x}}^m)],
\end{align}
where $P_C$ is the conditional distribution over styles deﬁned by the classifier.
%To improve the style transfer accuracy, we employ a pre-trained style classifier $C(\bm{x})$ to provide 
%on transferring style. %We set the architecture of the classifier network has the same structure as Encoder. 
%We use the cross-entropy loss $\mathcal{L}_{style}$ for training:
%\begin{gather}
%p(\bm{s}\mid \bm{x}) = {\rm softmax}({\rm C}(\bm{x})W + b),
%\mathcal{L}_{style} = -\log p(\bm{s}\mid \bm{x})
%\end{gather}
%where W and b are trainable parameters.
We train the classifier on the whole training set with the standard cross-entropy loss. Then, we freeze the weights of style classifier for computing $\mathcal{L}_{\rm style}$. On the other hand, %since the sampling process does not allow gradient back-propagation, 
we follow \citet{lee-etal-2021-enhancing,dai-etal-2019-style} to use soft sampling to allow gradient back-propagation. %Specifically, while the gradient ﬂow is returned from classifier to the decoder in Figure~\ref{training}, we leverage the product of probability distribution of each step and the weight of embedding layer to project the representation of outputs onto word embedding space.
%\paragraph{Learning Algorithm}
%As shown in Figure~\ref{training}, the learning objective function of can be represented as:

%where $\mathcal{L}_{self}$, $\mathcal{L}_{sen}$, $\mathcal{L}_{style}$ denote the self reconstruction loss, the sen loss, the style classifier loss, respectively.
% \iffalse
% \paragraph{Self Reconstruction Loss}As discussed in section 1 (\S(\ref{1})), decoder employs a left-to-right generator conditioned on stylized sentence-level representations. As for training, decoder aims to reconstruct input in an unsupervised environment through minimizing log-likelihood $\mathcal{L}_{self}$ human-written texts. Specifically, 
% \begin{gather}
% h_t = {\rm Decoder}(y^m_{<t},\{z_i^{\hat{s}}\}_i^{n+1})\\
% p(y_t^m\mid y_{<t}^m,\bm{x^m},\bm{\hat{s}})= {\rm softmax}(h_tW + b) \\
% \mathcal{L}_{self} = -\sum_{t=1}^j\log p(y_t^m\mid y_{<t}^m,\bm{x^m},\bm{\hat{s}})
% \end{gather}
% where $h_t$ is the decoder’s hidden state at the t-th position computed from the context (i.e., the preﬁx $y_{<t}$ and $\bm{z_{srt}}$), W and b are trainable parameters.
% \fi

\subsection{Content Preservation Enhancing}

% \subsection{Identifying Style-Specific Contents} \label{3.3}
As aforementioned, author styles have a strong correlation with contents. It is difficult to transfer such style-specific contents to other styles directly. Since we train the model in an auto-encoder manner, it will have no idea how to transfer those content representations that have never seen other style embeddings during training.
%into this style in the inference stage. Therefore, 
To address the issue,  we propose to mask the style-specific keywords in the source text and perform style transfer on the masked text in the first generation stage. Then, we fill the masked tokens in the second stage. %by on the generalizability of pre-trained language models, which reduces the difficulty of transfer.

%previous studies shown, a sound style transfer system could preserve the characters and main plots of a story. For plots, different authors may have different narrative styles, which can be transferred by sentence representation. However, some attributes (e.g., names) of characters are style-specific content which won't show up in another author's writing. To tackle the above problems, we extract and fill the style-specific content in the second stage, explicitly. 

%As for data pre-processing, we could not extract all style-specific content. Nevertheless, 
 We follow \citet{xiao2021transductive} to use a frequency-based method to identify the style-specific keywords. Specifically, we extract style-specific keywords by (1) obtaining the top-$10$ words with the highest TF-IDF scores from each corpus, (2) retaining only people's names, place names, and proper nouns, (3) and filtering out those words with a high frequency in all corpora\footnote{We set those words appearing in at least 10\% samples in a corpus as high-frequency words.}. We denote the resulting word set as $D^s$ for the corpus with the style $s$. We extract the style-specific keywords $\boldsymbol{k}$ from the text $\boldsymbol{x}$ by selecting the words that are in $D^{s}$.
%conduct the TF-IDF (term frequency–inverse document frequency) algorithm on each corpus to obtain the top-k words. Then we filter out the style unrelated words. Concretely, we regard the words with the high frequency in all styles corpus as style unrelated words, denoting as $\textit{D}$. 
% the  use the part-of-speech tagging toolkit to filter  in top-k words.
%\begin{gather}
%\{w_i\}_i^k= {\rm TF\text{-}IDF}(\bm{x}) \\
%\bm{k}=\{w_i| w_i\notin \textit{D}\}
%\end{gather}
We detail above operation 
and explain it 
in Appendix
~\ref{keywords}. 
% Where the function of $\rm TF\text{-}IDF$ indicates the TF-IDF algorithm, and we detail the operation of filter in Appendix~\ref{keywords}. 
% and $\rm Filter$ denotes the operation of filter. then utilize a part-of-speech tagging toolkit (e.g., NLTK~\cite{bird2009natural}) to refine the keywords (e.g., characters' name), which is the subset of the style-specific content.

In the second stage, 
we train another model to fill the mask tokens in outputs of the first stage conditioned on the identified style-specific keywords in source inputs. %Put simply, it is a word filling model based on encoder-decoder structure. To achieve this, 
During training, we concatenate the keywords in $\bm{k}$ with a special token $\langle\texttt{Key}\rangle$ and feed them into the encoder paired with $\bm{x}^m$, as shown in Figure~\ref{Fig_1}. The training object is formulated as Equation \ref{dae_loss}. During inference, the decoder generates the transferred text $\hat{\boldsymbol{x}}$ conditioned on the output of the first stage $\hat{\boldsymbol{x}}^m$ in an auto-regressive manner.

\section{Experiments}
\subsection{Datasets}

\begin{table}[!t]
    \footnotesize
    \centering
    \scalebox{0.95}{
    \begin{tabular}{cc|ccc|c|c}
    
    \toprule
    \multicolumn{2}{c|}{\textbf{Dataset}} & \multicolumn{3}{c|}{\textbf{Train}} & \textbf{Val}& \textbf{Test}
    \\
    % &\multicolumn{3}{c}{\textbf{Chinese}} & \multicolumn{3}{c}{\textbf{English}} \\
    % &\textbf{Train} & \textbf{Valid}  & \textbf{Test} &\textbf{Train} & \textbf{Valid}  & \textbf{Test} \\
    \midrule
    \multirow{3}{*}{\textbf{ZH}}
    & \textbf{Style} & JY & LX &Tale & Tale & Tale\\
    & \textbf{Size} & 2,964 & 3,036 & 1,456 & 242 & 729
    \\
    & \textbf{Avg Len} & 344 & 168 & 175 & 175 & 176 \\
    \midrule
    \multirow{2}{*}{\textbf{EN}}
    & \textbf{Style} & \multicolumn{2}{c}{Shakespeare} &  ROC &ROC &ROC\\
    & \textbf{Size} & \multicolumn{2}{c}{1,161}  & 1,161 & 290 &290 \\
    & \textbf{Avg Len} & \multicolumn{2}{c}{71}  & 49 & 48 & 50\\
  \bottomrule
    \end{tabular}}
    \caption{Statistics of the Chinese~(ZH) and English~(EN) datasets. Avg Len indicates the average length of tokens of each sample. } %the sizes and styles of train and test sets. 
    %The tale in Chinese and story in English indicate fairy tales from LOT \cite{guan2021lot} and everyday stories from ROCStories \cite{mostafazadeh-etal-2016-corpus}, respectively.}
    \label{data_sta}
\end{table}

We construct stylized story datasets in Chinese and English, respectively.
%Due to the lack of stylized long text datasets, we built two datasets in Chinese and English, respectively. 
The Chinese dataset consists of three styles of texts, including fairy tales from LOT~\cite{guan2021lot}, LuXun~(LX), and JinYong~(JY). Specifically, LuXun writes realism novels while JinYong focuses on martial arts novels. 
These texts of different styles have a gap in lexical, syntactic, and semantic levels. 
Samples of different styles are detailed in Appendix~\ref{sample_analyse}. 

In our experiments, we aim to transfer a fairy tale to the LX or JY style. The English dataset consists of two styles of texts, including everyday stories from ROCStories~\cite{mostafazadeh-etal-2016-corpus} and fragments from Shakespeare's plays. We expect to transfer a five-sentence everyday story into the Shakespeare style. The statistics of datasets are shown in Table~\ref{data_sta}. The more details are described in Appendix~\ref{data_proprecessing}.

\subsection{Implementation}
We take LongLM$_{\rm BASE}$~\cite{guan2021lot} and T5$_{\rm BASE}$~\cite{2020t5} as the backbone model of both generation stages for Chinese and English experiments, respectively. 
%We take Transformer$_{BASE}$ (220M) \cite{NIPS2017_3f5ee243} as the backbone of our model. 
% Specifically, both the encoder and decoder consist of 12 layers. 
Furthermore, the fusion module and pointer network consist of two and one layers of randomly initialized bidirectional Transformer blocks~\cite{NIPS2017_3f5ee243}, respectively. 
We conduct experiments on one RTX 6000 GPU. In addition, we build the style classifier based on the encoder of LongLM$_{\rm BASE}$ and T5$_{\rm BASE}$ for Chinese and English, respectively. 

We set $\lambda_1$/$\lambda_2$/$\lambda_3$ in Equation~\ref{stage1} to 1/1/1, the batch size to 4, the learning rate to 5e-5, the maximum sequence length of the encoder and decoder to 512 for both generation stages in the Chinese experiments. And the hyper-parameters for English experiments are the same except that $\lambda_1$/$\lambda_2$/$\lambda_3$ are set to 0.5/0.5/0.5 and the learning rate to 2.5e-5. More implementation details are presented in Appendix~\ref{more_implementation_details}.

\begin{table*}[!t]
    \scriptsize
    \centering
    \begin{tabular}{c|l|ccccccc|cc}
    \toprule
    \textbf{Target Styles}
    &\textbf{Models}
    & \textbf{r-Acc}&\textbf{a-Acc} & \textbf{BLEU-1} & \textbf{BLEU-2} & \textbf{BS-P} & \textbf{BS-R} & \textbf{BS-F1} & \textbf{BL-Overall} & \textbf{BS-Overall}

    \\
    \midrule

\multirow{4}{*}{\textbf{ZH-LX}}
&\textbf{Style Transformer}& \text{65.84} &\text{0.13} & \text{82.53} & \text{77.17} & \text{96.92} & \text{96.51} & \text{96.70} & \text{2.96} & \text{3.26} \\

&\textbf{StyleLM}& \text{97.80} &\text{33.33} & \text{39.43} & \text{19.66} & \text{77.71} & \text{75.02} & \text{76.30} & \text{31.38} & \text{50.42} \\

&\textbf{Reverse Attention}&\text{98.49} & \text{42.93} & \text{20.98} & \text{6.70} & \text{65.38} & \text{63.39} & \text{64.35} & \text{24.37} & \text{52.55} \\

    \cline{2-11}
&\textbf{StoryTrans} &\text{97.66} &\text{59.94} & \text{32.19} & \text{14.44} & \text{68.53} & \text{70.48} & \text{69.45} & \textbf{37.38} & \textbf{64.52} \\

% TODO:继续修改

\midrule

\multirow{4}{*}{\textbf{ZH-JY}}&\textbf{Style Transformer}& \text{46.77} &\text{0.13} & \text{83.24} & \text{77.85} & \text{97.15} & \text{96.82} & \text{96.97} & \text{3.23} & \text{3.55} \\

&\textbf{StyleLM}&\text{79.97}&\text{51.16} & \text{36.72} & \text{18.01} & \text{74.20} & \text{75.19} & \text{74.62} & \text{37.41} & \text{61.78}\\

&\textbf{Reverse Attention}&\text{94.51} &\text{66.39} & \text{21.15} & \text{6.32} & \text{64.05} & \text{65.08} & \text{64.54} & \text{30.19} & \text{65.45}\\
    \cline{2-11}
&\textbf{StoryTrans}&\text{84.49} & \text{62.96} & \text{30.71} & \text{14.5} & \text{68.76} & \text{71.69} & \text{70.16} & \textbf{37.72} & \textbf{66.46}\\

\midrule
\midrule

\multirow{4}{*}{\textbf{EN-SP}}&\textbf{Style Transformer}&\text{0.34}&\text{0.01} & \text{99.88} & \text{99.88} & \text{87.10} & \text{95.43} & \text{90.78} & \text{3.31} & \text{3.16}\\

&\textbf{StyleLM}&\text{57.93}&\text{3.44} & \text{37.05} & \text{19.40} & \text{84.72} & \text{90.53} & \text{87.30} & \text{9.85} & \text{17.32}\\

&\textbf{Reverse Attention}&\text{20.68}&\text{0.01} & \text{96.90} & \text{96.16} & \text{86.93} & \text{95.27} & \text{90.61} & \text{3.25} & \text{3.15}\\
    \cline{2-11}
&\textbf{StoryTrans}&\text{88.62} &\text{52.41} & \text{32.20} & \text{12.71} & \text{81.77} & \text{87.51} & \text{84.31} & \textbf{34.31} & \textbf{66.47}\\

  \bottomrule 
    \end{tabular}
    \caption{Automatic evaluation results on the test set of the Chinese and English datasets. Bold numbers indicate best performance. ZH-LX/ZH-JY is the Chinese author LuXun/JinYong, respectively. EN-SP is the English author Shakespeare. StoryTrans achieves the best overall performance~(BL/BS-Overall), with a good trad-off between style accuracy~(r/a-Acc) and content preservation~(BLEU-1/2 and BS-P/R/F1). 
    % MS and CE denote mutiple $<Sen>$ and content enhancing, respectively.
    }
    \label{results_zh}
\end{table*}

\begin{table*}[!t]
    \scriptsize
    \centering
    \begin{tabular}{l|ccccccc|cc}
    \toprule
    % \textbf{Target Styles}
    % \textbf{}
    & \textbf{r-Acc}&\textbf{a-Acc} & \textbf{BLEU-1} & \textbf{BLEU-2} & \textbf{BS-P} & \textbf{BS-R} & \textbf{BS-F1} & \textbf{BL-Overall} & \textbf{BS-Overall}

    \\
    \midrule

\textbf{Proposed Model}& \text{88.62} &\text{52.41} & \text{32.20} & \text{12.71} & \text{81.77} & \text{87.51} & \text{84.31} & \text{34.31} & \text{66.47} \\

\midrule

\textbf{(-)~$\mathcal{L}_{\rm dis}$}& \text{75.86} &\text{31.37} & \text{33.49} & \text{14.52} & \text{82.38} & \text{88.07} & \text{84.92} & \text{27.44} & \text{51.61} \\

\textbf{(-)~$\mathcal{L}_{\rm style}$}&\text{50.68} & \text{7.93} & \text{45.00} & \text{23.79} & \text{84.38} & \text{89.16} & \text{86.5} & \text{16.51} & \text{26.19} \\

\textbf{(-)~$\mathcal{L}_{\rm sop}$} &\text{78.96} &\text{38.96} & \text{39.45} & \text{19.20} & \text{82.92} & \text{88.62} & \text{85.47} & \text{33.80} & \text{57.70} \\

\textbf{(-)}~CE &\text{92.41} &\text{73.10} & \text{21.62} & \text{6.09} & \text{79.73} & \text{86.12} & \text{82.59} & \text{31.82} & \text{77.70} \\

  \bottomrule 
    \end{tabular}
    \caption{Ablation study results on English datasets.~(-) indicates removing the component in proposed model.  CE denote content enhancing, which means removing the second stage. More ablation results shown in Appendix~\ref{more_ablation}.
    }
    \label{en_ablation}
\end{table*}

\subsection{Baselines}
Since no previous studies have focused on story author-style transfer, we build several baselines by adapting short-text style transfer models. %Loosely speaking, we apply the state-of-the-art style transfer ideas to the long text generation model. 
For a fair comparison, we initialize all baselines using the same pre-trained parameters as our model.
%Since long text generation is also still a hard task, all baselines and our models are initialized from pre-trained model. 
Specifically, we adopt the following baselines:
\\
\textbf{Style Transformer:} %Following the setting of Style Transformer \cite{dai-etal-2019-style}, we employs style embedding to control style intensity and set several auxiliary losses to train on non-parallel data. 
It adds an extra style embedding and a discriminator to provide style transfer rewards without disentangling content from styles~\cite{dai-etal-2019-style}.\\
\textbf{StyleLM:} This baseline generates the target text conditioned on the given style token and corrupted version of the original text~\cite{syed2020adapting}.\\
\textbf{Reverse Attention:} It inserts a reverse attention module on the last layer of the encoder, which aims to negate the style information from the hidden states of the encoder~\cite{lee-etal-2021-enhancing}.

\subsection{Automatic Evaluation}
\paragraph{Evaluation Metrics}
Previous works evaluate style transfer systems mainly from three aspects including style transfer accuracy, content preservation, and sentence fluency.
% ~\cite{lee-etal-2021-enhancing, ijcai2019-732}.
%In style transfer tasks, it's a contradiction of content preservation and style accuracy. 
A good style transfer system needs to balance the contradiction between content preservation and transfer accuracy~\cite{zhu-etal-2021-neural,niu-bansal-2018-polite}. We use a joint metric to evaluate the overall performance of models. On the other hand, previous studies usually use perplexity~(PPL) of a pre-trained language model. However, in our experiments, we found that the PPL of model outputs is lower than human-written texts, suggesting that PPL is not reliable for evaluating the quality of stories. %This may be caused by stylized long text, which is hard to evaluate ﬂuency by pretrained model based on PPL. 
Therefore, we evaluate the fluency through manual evaluation.
%We  the additional coherence in human evaluation and give up the fluency evaluated by PPL in automatic evaluation. We will investigate this issue in future work.

Specifically, we adopt the following automatic metrics: \textbf{(1) {Style Transfer Accuracy}:} 
We use two variants of style transfer accuracy following \citet{krishna2021few}, absolute accuracy~(a-Acc) and relative accuracy~(r-Acc). We train a style classifier and regard the classifier score as the a-Acc.
% ~\cite{lee-etal-2021-enhancing,ijcai2020-526}. 
And r-Acc is a binary value to indicate whether the style classifier score the output higher than the input~(1/0 for a higher/lower score). We train the classifier by fine-tuning the encoder of LongLM$_{\rm BASE}$ and T5$_{\rm BASE}$ on the Chinese and English training set, respectively. %After pretraining, the parameter of style classifier are frozen. 
The classifier achieves a 99.6\% and 99.41\% accuracy on the Chinese and English test sets, respectively. \textbf{(2) {Content Preservation}:} We use BLEU-$n$~($n$=1,2)~\cite{10.3115/1073083.1073135} and BERTScore~(BS)~\cite{bert-score} between generated and input texts to measure their lexical and semantic similarity, respectively. And we report recall~(BS-R), precision~(BS-P) and F1 score~(BS-F1) for BS. \textbf{(3) {Overall}:} 
We use the geometric mean of a-ACC and BLEU/BS-F1 score~(BL-Overall/BS-Overall) to assess the overall performance of models \cite{krishna-etal-2020-reformulating,lee-etal-2021-enhancing}. %Specifically,
%\begin{gather}
    %J(x) = \sqrt{Sty(x)\cdot Con(x)}
%\end{gather}
%where $J(x)$ denotes the J-Score. And in this case, we employs Acc and BLEU-1 as $Sty(x)$ and $Con(x)$, respectively.

\paragraph{Results on the Chinese Dataset} %We compare LongTrans with baselines on the test sets, and show the automatic evaluation results in Table~\ref{results}. %LongTrans achieves the best overall performance on both Chinese and English datasets, which indicates LongTrans possesses a more robust style transfer ability on both Chinese and English.
We show the overall performance and individual metrics results in Table~\ref{results_zh}. In terms of overall performance, StoryTrans outperforms baselines, illustrating that StoryTrans can achieve a better balance between style transfer and content preservation.

In terms of style accuracy, 
% as shown in 
% Table~\ref{results_zh}, %by the results on the Chinese dataset, 
StoryTrans achieves the best style transfer accuracy~(a-Acc) in LX and comparable performance in JY. 
The bad performance of baselines indicates the necessity to perform explicit disentanglement 
% between styles and contents 
beyond the token level.
% for story style transfer. 
% Otherwise, the style-specific contents XXX.
%Reverse Attention gets , This indicates that disentanglement on token-level representation is insufficient. Compared with LongTrans, it is more suitable to apply disentanglement on high-level representations for long text style transfer. 
In addition, manual inspection shows that Style Transformer tends to copy the input, accounting for the highest BLEU score and BERTScore. This means Style Transformer only takes the target style signals as noise, which may result from the stylistic features existing in the contents.
StyleLM and Reverse Attention get better transfer accuracy than Style Transformer by removing such stylistic features from the contents. Moreover, Reverse Attention obtains better style accuracy but worse content preservation than StyleLM. Therefore, re-weighting hidden states 
allows better control over style than deleting input words explicitly.

In terms of content preservation, StoryTrans outperforms Reverse Attention. Additionally, StyleLM achieves better performance in content preservation, benefiting from inputting noisy versions of golden texts. But without disentanglement, it can't strip style information. This leads to a lower overall performance than StoryTrans. As for Style Transformer, the results demonstrate that only an attention-based model hardly removes style features in overwhelming tokens information, leading to degenerate into an auto-encoder.

\begin{figure*}[t!]
\centering
\includegraphics[width=1\linewidth]{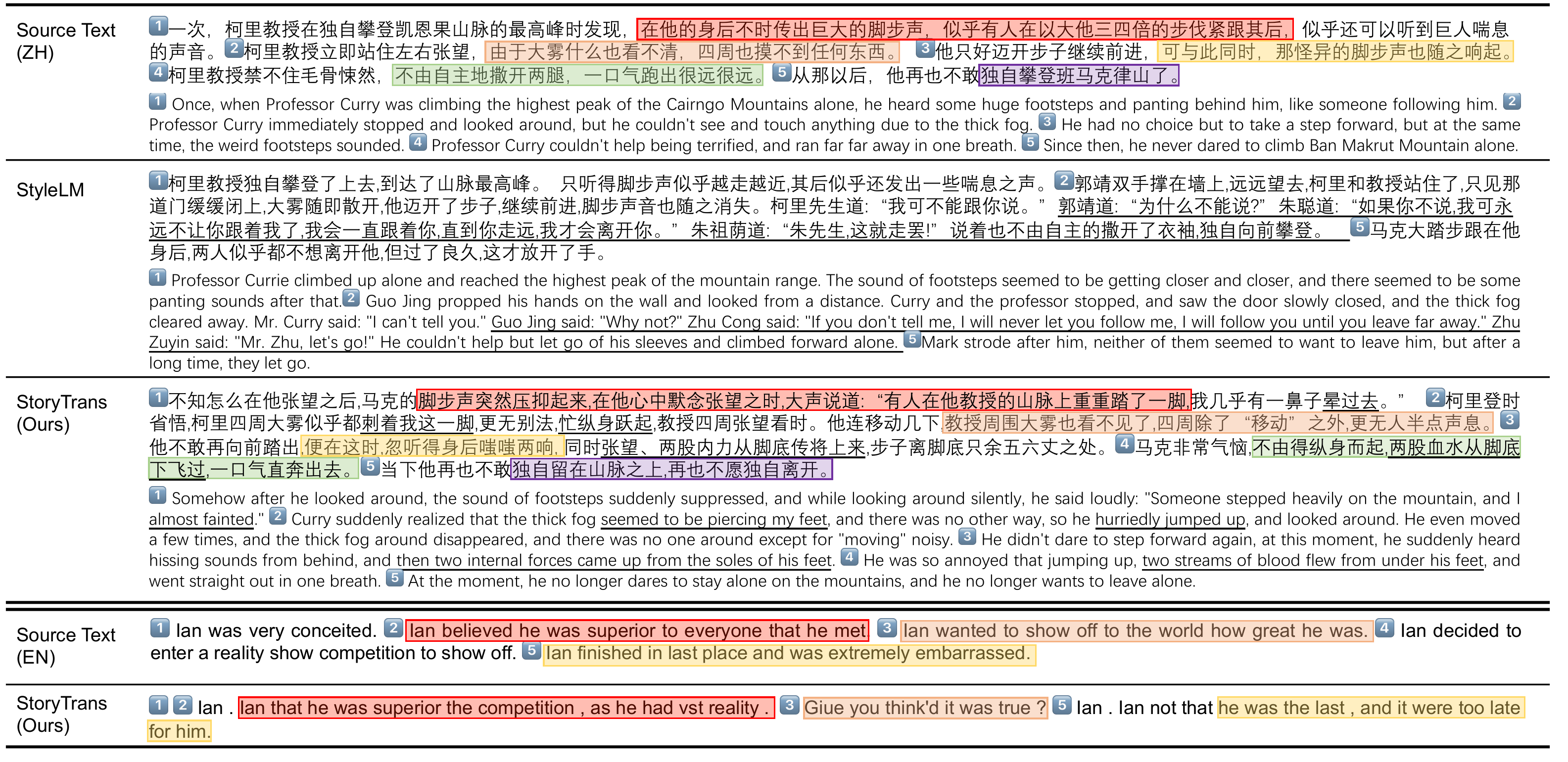}
\captionsetup{type=table}
\caption{Cases generated by different models, which are transferred from the fairy tale style~(ZH) to the JY style and every-day story~(EN) to Shakespeare style, respectively. The number before each sentence in the generated cases is the corresponding sentence in the source text in semantics. The underlined sentences or short phrase indicate inserted contents to align with the target style. We highlight the rewritten contents in corresponding colors between the source and generated texts. The English texts below the Chinese are translated versions of the Chinese samples. More case shown in Appendix~\ref{more_case_study}.}
\label{case_study}
\end{figure*}

\begin{table}[t!]
    \scriptsize
    \centering
    \scalebox{0.7}{
    \begin{tabular}{l|llll|llll}
    \toprule
    \multirow{2}{*}{\textbf{Models}}&\multicolumn{4}{c|}{\textbf{LX}} 
    & \multicolumn{4}{c}{\textbf{JY}}
    \\
    & \textbf{Sty.} & \textbf{Con.} & \textbf{Coh.} & $\kappa$
    & \textbf{Sty.} & \textbf{Con.} & \textbf{Coh.} & $\kappa$
    \\
    \midrule
    \textbf{Style Transformer} & \text{1.02} & \textbf{2.95**} & \textbf{2.91**} & 0.80
    & \text{1.00} & \textbf{2.98**} & \textbf{2.94**} & 0.89
    \\
    \textbf{StyleLM}  & \text{1.61} & \text{1.99} & \text{1.58} & 0.20
    & \text{1.7} & \text{1.92} & \text{1.94} & 0.23
    \\
    \textbf{Reverse Attention} & \text{1.69} & \text{1.25} & \text{1.64} & 0.21 & \text{2.07} & \text{1.25} & \text{1.92} & 0.20
    \\
    \midrule
    \textbf{StoryTrans} & \textbf{1.98**} & \text{1.84} & \text{1.67} & 0.24
    & \textbf{2.43**} & \text{1.69} & \text{1.91} & 0.23
    \\

  \bottomrule
    \end{tabular}}
    \caption{Human evaluation results on Chinese for transfer direction in LX and JY. $\kappa$ denotes Fleiss’ kappa \cite{fleiss1971measuring} to measure the inter annotator agreement (all are moderate or substantial). The scores marked with $\ast\ast$ mean StoryTrans outperforms the baselines signiﬁcantly with p-value<0.01 (sign test).}
    \label{human_eval}
\end{table}

\begin{figure}[t!]
\centering
\includegraphics[width=0.9\linewidth]{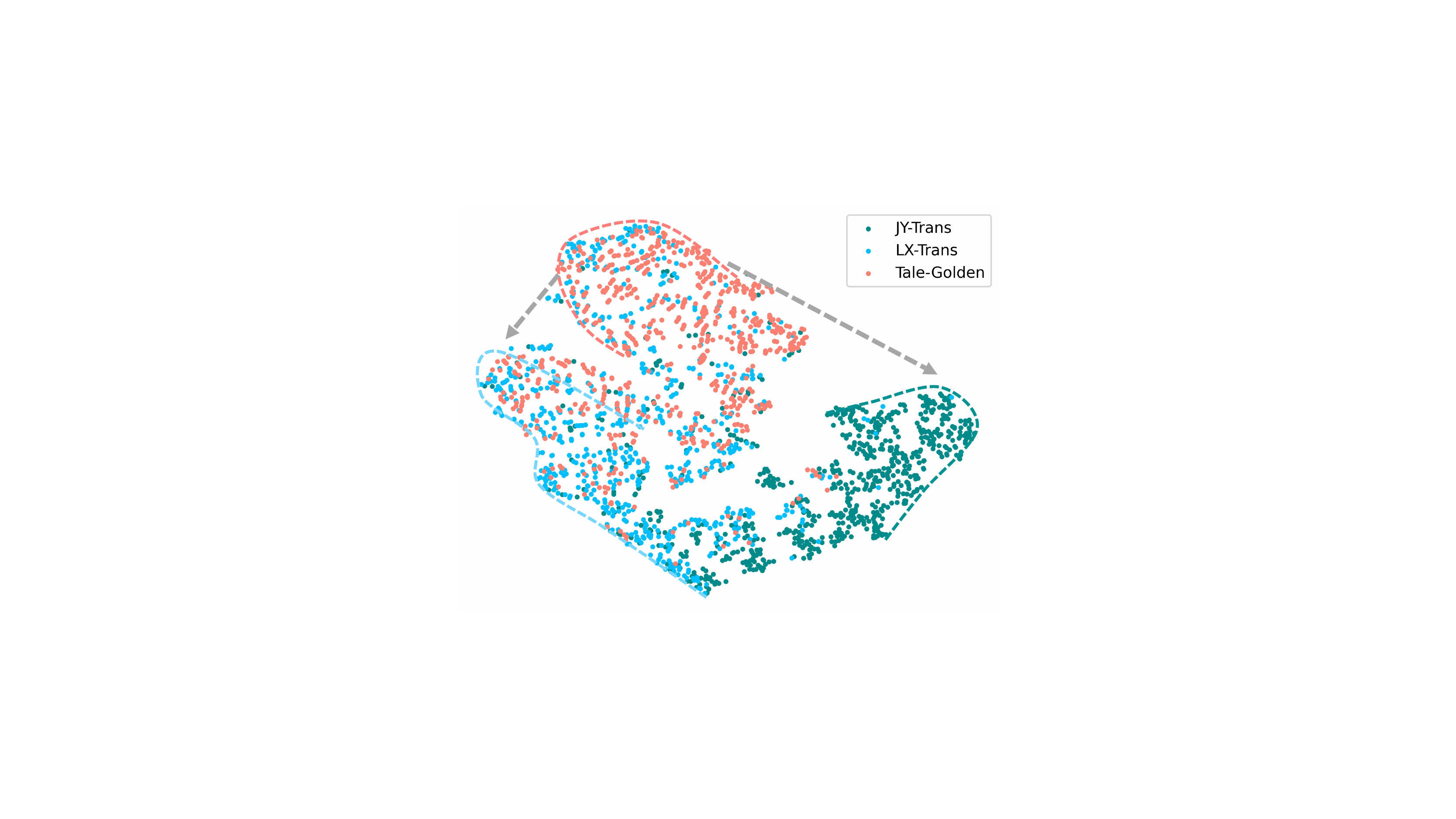}
\caption{Stylistic features visualization of the golden texts~(-Golden) and generated texts~(-Trans) on the Chinese test set using t-SNE~\cite{hinton2002stochastic}.} 
\label{style_analysis}
\end{figure}

\paragraph{Results on the English Dataset} %In order to validate the ability of LongTrans, we perform the same experiments on English datasets and results also presented in Table \ref{results}. 
Similarly, StoryTrans achieves the best overall performance on the English dataset, showing its effectiveness and generalization. And StoryTrans outperforms baselines significantly in terms of style transfer accuracy. %indicating that baselines' ability of style control is poorer. 
As for content preservation, Style Transformer and Reverse Attention degenerate into an auto-encoder, and tend to copy the input even more than their performance on the Chinese dataset. 

\paragraph{Results on Ablation Study}

As shown in Table~\ref{en_ablation}, we observe a significant drop in transfer accuracy without $\mathcal{L}_{\rm dis}$ or $\mathcal{L}_{\rm style}$. %suggesting the necessity of disentanglement and style supervision for self-reconstruction. 
$\mathcal{L}_{\rm dis}$ works by disentangling stylistic features from the discourse representations, while $\mathcal{L}_{\rm style}$ exerts direct supervision on styles of generated texts. Without $\mathcal{L}_{\rm sop}$, model can hardly capture discourse-level information and keeps more source tokens, leading to higher BLEU scores and lower accuracy.
When removing the second stage, the lowers BLEU scores show the benefit of the mask-and-fill framework for content preservation. 

\subsection{Manual Evaluation}
We randomly sampled 100 fairy tales from the Chinese dataset, and obtained 800 generated texts 
% for two target styles 
from StoryTrans and three baseline models. %Given the source texts and the target style, 
Then, we hire three Chinese native speakers 
% as annotators 
to evaluate 
% generated texts 
in 
% the following 
three aspects including style transfer accuracy (\textbf{Sty.}), content preservation (\textbf{Con.}) and coherence (\textbf{Coh.}). %As mentioned before, we introduce 
% and grammatical correctness, and inter-sentence causal and temporal dependencies. 
We ask the annotators to judge each aspect from 1~(the worst) to 3~(the best).
%evaluate generated texts with a score ranging from 1(worst) to 3(best) in each aspect. 
% And we compute the final score of each text by averaging the scores of three annotators. %Table~\ref{human_eval}. 
%Since we could not find annotators who are familiar with Shakespeare's text, human evaluation performs only on the Chinese dataset.

As illustrated in Table \ref{human_eval}, our StoryTrans received the highest style accuracy and modest performance in content preservation and coherence. More details and analysis are presented in Appendix~\ref{ana_manual_eva}.

\subsection{Case Study}

Table~\ref{case_study} shows the cases generated by StoryTrans and the best baseline.
StyleLM inserts many unrelated sentences, which overwhelm the original content and impact the coherence, further leading to the content loss of sentences 3 and 4. 
On the contrary, StoryTrans supplement several short phrases or plots~(e.g., \begin{CJK*}{UTF8}{gkai}{“纵身跃起”}\end{CJK*}~/“hurriedly jumped up”) to enrich the storyline and maintain the main content. Furthermore, StoryTrans can rewrite most sentences with the target style and maintain source semantics. 
In addition, StyleLM tends to discard the source entities and use words which is specific in the target style~(e.g., \begin{CJK*}{UTF8}{gkai}{“郭靖”}\end{CJK*}~/“Guo Jing”), while StoryTrans dose not, suggesting the necessity of the mask-and-fill framework.
% We show more cases in Appendix~\ref{more_case_study}.

\subsection{Stylistic Feature Visualization}
% In order to analyze how model transfer the style, 
We follow \citet{syed2020adapting} to define several stylistic features and visualize the features of the golden texts and generated texts on the Chinese test set. The stylistic features include the type and number of punctuation marks, the number of sentences, and the number of words. As shown in Figure~\ref{style_analysis}, the texts generated by Reverse Attention and StyleLM have similar stylistic features to source texts. In contrast, StoryTrans can better capture different stylistic features and transfer source texts to specified styles.
%visualize the golden Chinese test dataset and transferred texts in JY and LX styles projected in style space using t-SNE in Figure~\ref{style_analysis}. 
% Specifically, since features of the author style are more abstract, we followed \citet{syed2020adapting} to adopt surface elements of texts to show the author style. The surface element is an indicative notion revealing author’s writing style like expression. 
%Specifically, we count surface elements~\cite{syed2020adapting} of texts into feature vectors to represent the style. Surface elements include the type and number of punctuation marks, the number of sentences and the number of words.
%As shown in Figure\ref{style_analysis}, we can clearly observe that each style texts are separated. 
% while they are similar in semantics. 
%This ﬁgure illustrates that our LongTrans imitate target author writing style to transfer the texts. 
More details are in Appendix~\ref{apd:style_analaysis}.

\section{Conclusion}

In this paper, we present the first study for story author-style transfer and analyze the difficulties of this task. Accordingly, we propose a novel generation model, which explicitly disentangles the style information from high-level text representations to improve the style transfer accuracy, and achieve better content preservation by injecting style-specific contents.
%explore a challenging long text author-style transfer task and propose a novel framework called LongTrans to transferred vernacular texts into interesting stylized novels. 
%LongTrans leverages disentanglement loss to reduce style attributes on discourse representation, and then combines the style embedding to generate target author-style texts. LongTrans also use mask-and-fill two-stage framework to enhance content preservation. 
Automatic evaluations show StoryTrans outperform baselines on the overall performance. Further analysis shows StoryTrans has a better ability to capture linguistic features for style transfer.
% Further analysis shows that LongTrans has better ability of style controlling thanks to disentangling on discourse representation. 
% In the future, we will investigate how to generate more fluency while keeping high-level style accuracy and content preservation. 

% Entries for the entire Anthology, followed by custom entries

\section{Limitations}
In style transfer, content preservation and style transfer are adversarial. Long texts have richer contents and more abstract stylistic features. We also notice that content preservation is the main disadvantage of StoryTrans in automatics evaluation results. Case studies also indicate that StoryTrans can maintain some entities and the relations between entities. However, strong discourse-level style transfer ability endangered content preservation. In contrast, baselines such as Style Transformer have better content preservation but hardly transfer the style. We believe that StoryTrans is still a good starting point for this important and challenging task.

During preliminary experiments, we also manually inspected multiple author styles besides Shakespeare, such as Mark Twain. However, we found that their styles are not as obvious as Shakespeare, as shown in the following example. Therefore, we only selected authors with relatively distinct personal styles for our transfer experiments. In future work, we will expand our research and choose more authors with distinct styles for style transfer. For example, the style distinction between the following examples is not readily apparent.
\begin{itemize}
    \item Everyday story in our datatset: \textit{Ashley wanted to be a unicorn for Halloween. She looked all over for a unicorn costume. She wasn't able to find one}.
    \item "A Double Barrelled Detective Story" by Mark Twain: \textit{You will go and find him. I have known his hiding-place for eleven years; it cost me five years and more of inquiry}.
\end{itemize}

\section{Acknowledgments}
This work was supported by the NSFC projects (Key project with No. 61936010 ). This work was also supported by the Guoqiang Institute of Tsinghua University, with Grant No. 2020GQG0005.

\section{Ethics Statement}
We perform English and Chinese experiments on public datasets and corpora. 
Specifically, English datasets come from ROCstories and Project Gutenberg. Moreover, Chinese datasets include the LOT dataset and public corpora of JY and LX.
Automatic and manual evaluation demonstrate that our model outperforms strong baselines on both Chinese and English datasets.
In addition, our model can be easily applied to different languages by substituting specific pre-trained language models.

As for manual evaluation, we hired three native Chinese speakers as annotators to evaluate generated texts and did not ask about personal privacy or collect the personal information of annotators. We pay 1.8 yuan (RMB) per sample in compliance with Chinese wage standards. Considering it would cost an average of 1 minute for an annotator to score a sample, the payment is reasonable.

% Entries for the entire Anthology, followed by custom entries
% \bibliography{anthology,custom}
\bibliography{custom}

\begin{thebibliography}{38}
\expandafter\ifx\csname natexlab\endcsname\relax\def\natexlab#1{#1}\fi

\bibitem[{Bahdanau et~al.(2015)Bahdanau, Cho, and Bengio}]{bahdanau2015neural}
Dzmitry Bahdanau, Kyung~Hyun Cho, and Yoshua Bengio. 2015.
\newblock Neural machine translation by jointly learning to align and
  translate.
\newblock In \emph{3rd International Conference on Learning Representations,
  ICLR 2015}.

\bibitem[{Bird et~al.(2009)Bird, Klein, and Loper}]{bird2009natural}
Steven Bird, Ewan Klein, and Edward Loper. 2009.
\newblock \emph{Natural language processing with Python: analyzing text with
  the natural language toolkit}.
\newblock " O'Reilly Media, Inc.".

\bibitem[{Dai et~al.(2019)Dai, Liang, Qiu, and Huang}]{dai-etal-2019-style}
Ning Dai, Jianze Liang, Xipeng Qiu, and Xuanjing Huang. 2019.
\newblock \href {https://doi.org/10.18653/v1/P19-1601} {Style transformer:
  Unpaired text style transfer without disentangled latent representation}.
\newblock In \emph{Proceedings of the 57th Annual Meeting of the Association
  for Computational Linguistics}, pages 5997--6007, Florence, Italy.
  Association for Computational Linguistics.

\bibitem[{Devlin et~al.(2019)Devlin, Chang, Lee, and
  Toutanova}]{devlin-etal-2019-bert}
Jacob Devlin, Ming-Wei Chang, Kenton Lee, and Kristina Toutanova. 2019.
\newblock \href {https://doi.org/10.18653/v1/N19-1423} {{BERT}: Pre-training of
  deep bidirectional transformers for language understanding}.
\newblock In \emph{Proceedings of the 2019 Conference of the North {A}merican
  Chapter of the Association for Computational Linguistics: Human Language
  Technologies, Volume 1 (Long and Short Papers)}, pages 4171--4186,
  Minneapolis, Minnesota. Association for Computational Linguistics.

\bibitem[{Fan et~al.(2018)Fan, Lewis, and Dauphin}]{fan-etal-2018-hierarchical}
Angela Fan, Mike Lewis, and Yann Dauphin. 2018.
\newblock \href {https://doi.org/10.18653/v1/P18-1082} {Hierarchical neural
  story generation}.
\newblock In \emph{Proceedings of the 56th Annual Meeting of the Association
  for Computational Linguistics (Volume 1: Long Papers)}, pages 889--898,
  Melbourne, Australia. Association for Computational Linguistics.

\bibitem[{Fleiss(1971)}]{fleiss1971measuring}
Joseph~L Fleiss. 1971.
\newblock Measuring nominal scale agreement among many raters.
\newblock \emph{Psychological bulletin}, 76(5):378.

\bibitem[{Gao et~al.(2019)Gao, Zhang, Lee, Galley, Brockett, Gao, and
  Dolan}]{gao-etal-2019-structuring}
Xiang Gao, Yizhe Zhang, Sungjin Lee, Michel Galley, Chris Brockett, Jianfeng
  Gao, and Bill Dolan. 2019.
\newblock \href {https://doi.org/10.18653/v1/D19-1190} {Structuring latent
  spaces for stylized response generation}.
\newblock In \emph{Proceedings of the 2019 Conference on Empirical Methods in
  Natural Language Processing and the 9th International Joint Conference on
  Natural Language Processing (EMNLP-IJCNLP)}, pages 1814--1823, Hong Kong,
  China. Association for Computational Linguistics.

\bibitem[{Gong et~al.(2016)Gong, Chen, Qiu, and Huang}]{gong2016end}
Jingjing Gong, Xinchi Chen, Xipeng Qiu, and Xuanjing Huang. 2016.
\newblock End-to-end neural sentence ordering using pointer network.
\newblock \emph{arXiv preprint arXiv:1611.04953}.

\bibitem[{Guan et~al.(2021{\natexlab{a}})Guan, Feng, Chen, He, Mao, Fan, and
  Huang}]{guan2021lot}
Jian Guan, Zhuoer Feng, Yamei Chen, Ruilin He, Xiaoxi Mao, Changjie Fan, and
  Minlie Huang. 2021{\natexlab{a}}.
\newblock \href {http://arxiv.org/abs/2108.12960} {Lot: A benchmark for
  evaluating chinese long text understanding and generation}.

\bibitem[{Guan et~al.(2021{\natexlab{b}})Guan, Mao, Fan, Liu, Ding, and
  Huang}]{guan-etal-2021-long}
Jian Guan, Xiaoxi Mao, Changjie Fan, Zitao Liu, Wenbiao Ding, and Minlie Huang.
  2021{\natexlab{b}}.
\newblock \href {https://doi.org/10.18653/v1/2021.acl-long.499} {Long text
  generation by modeling sentence-level and discourse-level coherence}.
\newblock In \emph{Proceedings of the 59th Annual Meeting of the Association
  for Computational Linguistics and the 11th International Joint Conference on
  Natural Language Processing (Volume 1: Long Papers)}, pages 6379--6393,
  Online. Association for Computational Linguistics.

\bibitem[{Hinton and Roweis(2002)}]{hinton2002stochastic}
Geoffrey~E Hinton and Sam Roweis. 2002.
\newblock Stochastic neighbor embedding.
\newblock \emph{Advances in neural information processing systems}, 15.

\bibitem[{Huang et~al.(2021)Huang, Chen, Wu, Guo, Zhu, and
  Huang}]{huang-etal-2021-nast}
Fei Huang, Zikai Chen, Chen~Henry Wu, Qihan Guo, Xiaoyan Zhu, and Minlie Huang.
  2021.
\newblock \href {https://doi.org/10.18653/v1/2021.findings-acl.138} {{NAST}: A
  non-autoregressive generator with word alignment for unsupervised text style
  transfer}.
\newblock In \emph{Findings of the Association for Computational Linguistics:
  ACL-IJCNLP 2021}, pages 1577--1590, Online. Association for Computational
  Linguistics.

\bibitem[{Jain et~al.(2019)Jain, Mishra, Azad, and
  Sankaranarayanan}]{jain2019unsupervised}
Parag Jain, Abhijit Mishra, Amar~Prakash Azad, and Karthik Sankaranarayanan.
  2019.
\newblock Unsupervised controllable text formalization.
\newblock In \emph{Proceedings of the AAAI Conference on Artificial
  Intelligence}, volume~33, pages 6554--6561.

\bibitem[{John et~al.(2019)John, Mou, Bahuleyan, and
  Vechtomova}]{john-etal-2019-disentangled}
Vineet John, Lili Mou, Hareesh Bahuleyan, and Olga Vechtomova. 2019.
\newblock \href {https://doi.org/10.18653/v1/P19-1041} {Disentangled
  representation learning for non-parallel text style transfer}.
\newblock In \emph{Proceedings of the 57th Annual Meeting of the Association
  for Computational Linguistics}, pages 424--434, Florence, Italy. Association
  for Computational Linguistics.

\bibitem[{Krishna et~al.(2021)Krishna, Nathani, Garcia, Samanta, and
  Talukdar}]{krishna2021few}
Kalpesh Krishna, Deepak Nathani, Xavier Garcia, Bidisha Samanta, and Partha
  Talukdar. 2021.
\newblock Few-shot controllable style transfer for low-resource settings: A
  study in indian languages.
\newblock \emph{arXiv preprint arXiv:2110.07385}.

\bibitem[{Krishna et~al.(2020)Krishna, Wieting, and
  Iyyer}]{krishna-etal-2020-reformulating}
Kalpesh Krishna, John Wieting, and Mohit Iyyer. 2020.
\newblock \href {https://doi.org/10.18653/v1/2020.emnlp-main.55} {Reformulating
  unsupervised style transfer as paraphrase generation}.
\newblock In \emph{Proceedings of the 2020 Conference on Empirical Methods in
  Natural Language Processing (EMNLP)}, pages 737--762, Online. Association for
  Computational Linguistics.

\bibitem[{Lee et~al.(2021)Lee, Tian, Xue, and Zhang}]{lee-etal-2021-enhancing}
Dongkyu Lee, Zhiliang Tian, Lanqing Xue, and Nevin~L. Zhang. 2021.
\newblock \href {https://doi.org/10.18653/v1/2021.acl-long.8} {Enhancing
  content preservation in text style transfer using reverse attention and
  conditional layer normalization}.
\newblock In \emph{Proceedings of the 59th Annual Meeting of the Association
  for Computational Linguistics and the 11th International Joint Conference on
  Natural Language Processing (Volume 1: Long Papers)}, pages 93--102, Online.
  Association for Computational Linguistics.

\bibitem[{Lee et~al.(2020)Lee, Hudson, Lee, and Manning}]{lee-etal-2020-slm}
Haejun Lee, Drew~A. Hudson, Kangwook Lee, and Christopher~D. Manning. 2020.
\newblock \href {https://doi.org/10.18653/v1/2020.emnlp-main.120} {{SLM}:
  Learning a discourse language representation with sentence unshuffling}.
\newblock In \emph{Proceedings of the 2020 Conference on Empirical Methods in
  Natural Language Processing (EMNLP)}, pages 1551--1562, Online. Association
  for Computational Linguistics.

\bibitem[{Li et~al.(2015)Li, Luong, and Jurafsky}]{li-etal-2015-hierarchical}
Jiwei Li, Thang Luong, and Dan Jurafsky. 2015.
\newblock \href {https://doi.org/10.3115/v1/P15-1107} {A hierarchical neural
  autoencoder for paragraphs and documents}.
\newblock In \emph{Proceedings of the 53rd Annual Meeting of the Association
  for Computational Linguistics and the 7th International Joint Conference on
  Natural Language Processing (Volume 1: Long Papers)}, pages 1106--1115,
  Beijing, China. Association for Computational Linguistics.

\bibitem[{Logeswaran et~al.(2018)Logeswaran, Lee, and
  Radev}]{logeswaran2018sentence}
Lajanugen Logeswaran, Honglak Lee, and Dragomir Radev. 2018.
\newblock Sentence ordering and coherence modeling using recurrent neural
  networks.
\newblock In \emph{Thirty-second aaai conference on artificial intelligence}.

\bibitem[{Mostafazadeh et~al.(2016)Mostafazadeh, Chambers, He, Parikh, Batra,
  Vanderwende, Kohli, and Allen}]{mostafazadeh-etal-2016-corpus}
Nasrin Mostafazadeh, Nathanael Chambers, Xiaodong He, Devi Parikh, Dhruv Batra,
  Lucy Vanderwende, Pushmeet Kohli, and James Allen. 2016.
\newblock \href {https://doi.org/10.18653/v1/N16-1098} {A corpus and cloze
  evaluation for deeper understanding of commonsense stories}.
\newblock In \emph{Proceedings of the 2016 Conference of the North {A}merican
  Chapter of the Association for Computational Linguistics: Human Language
  Technologies}, pages 839--849, San Diego, California. Association for
  Computational Linguistics.

\bibitem[{Niu and Bansal(2018)}]{niu-bansal-2018-polite}
Tong Niu and Mohit Bansal. 2018.
\newblock \href {https://doi.org/10.1162/tacl_a_00027} {Polite dialogue
  generation without parallel data}.
\newblock \emph{Transactions of the Association for Computational Linguistics},
  6:373--389.

\bibitem[{Papineni et~al.(2002)Papineni, Roukos, Ward, and
  Zhu}]{10.3115/1073083.1073135}
Kishore Papineni, Salim Roukos, Todd Ward, and Wei-Jing Zhu. 2002.
\newblock \href {https://doi.org/10.3115/1073083.1073135} {Bleu: A method for
  automatic evaluation of machine translation}.
\newblock In \emph{Proceedings of the 40th Annual Meeting on Association for
  Computational Linguistics}, ACL '02, page 311–318, USA. Association for
  Computational Linguistics.

\bibitem[{Raffel et~al.(2020)Raffel, Shazeer, Roberts, Lee, Narang, Matena,
  Zhou, Li, and Liu}]{2020t5}
Colin Raffel, Noam Shazeer, Adam Roberts, Katherine Lee, Sharan Narang, Michael
  Matena, Yanqi Zhou, Wei Li, and Peter~J. Liu. 2020.
\newblock \href {http://jmlr.org/papers/v21/20-074.html} {Exploring the limits
  of transfer learning with a unified text-to-text transformer}.
\newblock \emph{Journal of Machine Learning Research}, 21(140):1--67.

\bibitem[{Reimers and Gurevych(2019)}]{reimers-2019-sentence-bert}
Nils Reimers and Iryna Gurevych. 2019.
\newblock \href {https://arxiv.org/abs/1908.10084} {Sentence-bert: Sentence
  embeddings using siamese bert-networks}.
\newblock In \emph{Proceedings of the 2019 Conference on Empirical Methods in
  Natural Language Processing}. Association for Computational Linguistics.

\bibitem[{Shen et~al.(2017)Shen, Lei, Barzilay, and
  Jaakkola}]{10.5555/3295222.3295427}
Tianxiao Shen, Tao Lei, Regina Barzilay, and Tommi Jaakkola. 2017.
\newblock Style transfer from non-parallel text by cross-alignment.
\newblock In \emph{Proceedings of the 31st International Conference on Neural
  Information Processing Systems}, NIPS'17, page 6833–6844, Red Hook, NY,
  USA. Curran Associates Inc.

\bibitem[{Syed et~al.(2020)Syed, Verma, Srinivasan, Natarajan, and
  Varma}]{syed2020adapting}
Bakhtiyar Syed, Gaurav Verma, Balaji~Vasan Srinivasan, Anandhavelu Natarajan,
  and Vasudeva Varma. 2020.
\newblock Adapting language models for non-parallel author-stylized rewriting.
\newblock In \emph{Proceedings of the AAAI Conference on Artificial
  Intelligence}, volume~34, pages 9008--9015.

\bibitem[{Tan et~al.(2021)Tan, Yang, Al-Shedivat, Xing, and
  Hu}]{tan-etal-2021-progressive}
Bowen Tan, Zichao Yang, Maruan Al-Shedivat, Eric Xing, and Zhiting Hu. 2021.
\newblock \href {https://doi.org/10.18653/v1/2021.naacl-main.341} {Progressive
  generation of long text with pretrained language models}.
\newblock In \emph{Proceedings of the 2021 Conference of the North American
  Chapter of the Association for Computational Linguistics: Human Language
  Technologies}, pages 4313--4324, Online. Association for Computational
  Linguistics.

\bibitem[{Tikhonov and Yamshchikov(2018)}]{tikhonov2018guess}
Alexey Tikhonov and Ivan~P Yamshchikov. 2018.
\newblock Guess who? multilingual approach for the automated generation of
  author-stylized poetry.
\newblock In \emph{2018 IEEE Spoken Language Technology Workshop (SLT)}, pages
  787--794. IEEE.

\bibitem[{Vaswani et~al.(2017)Vaswani, Shazeer, Parmar, Uszkoreit, Jones,
  Gomez, Kaiser, and Polosukhin}]{NIPS2017_3f5ee243}
Ashish Vaswani, Noam Shazeer, Niki Parmar, Jakob Uszkoreit, Llion Jones,
  Aidan~N Gomez, \L~ukasz Kaiser, and Illia Polosukhin. 2017.
\newblock \href
  {https://proceedings.neurips.cc/paper/2017/file/3f5ee243547dee91fbd053c1c4a845aa-Paper.pdf}
  {Attention is all you need}.
\newblock In \emph{Advances in Neural Information Processing Systems},
  volume~30. Curran Associates, Inc.

\bibitem[{Wu et~al.(2019)Wu, Zhang, Zang, Han, and Hu}]{ijcai2019-732}
Xing Wu, Tao Zhang, Liangjun Zang, Jizhong Han, and Songlin Hu. 2019.
\newblock \href {https://doi.org/10.24963/ijcai.2019/732} {Mask and infill:
  Applying masked language model for sentiment transfer}.
\newblock In \emph{Proceedings of the Twenty-Eighth International Joint
  Conference on Artificial Intelligence, {IJCAI-19}}, pages 5271--5277.
  International Joint Conferences on Artificial Intelligence Organization.

\bibitem[{Xiao et~al.(2021)Xiao, Pang, Lan, Wang, Shen, and
  Cheng}]{xiao2021transductive}
Fei Xiao, Liang Pang, Yanyan Lan, Yan Wang, Huawei Shen, and Xueqi Cheng. 2021.
\newblock Transductive learning for unsupervised text style transfer.
\newblock \emph{arXiv preprint arXiv:2109.07812}.

\bibitem[{Xu et~al.(2018)Xu, Sun, Zeng, Zhang, Ren, Wang, and
  Li}]{xu-etal-2018-unpaired}
Jingjing Xu, Xu~Sun, Qi~Zeng, Xiaodong Zhang, Xuancheng Ren, Houfeng Wang, and
  Wenjie Li. 2018.
\newblock \href {https://doi.org/10.18653/v1/P18-1090} {Unpaired
  sentiment-to-sentiment translation: A cycled reinforcement learning
  approach}.
\newblock In \emph{Proceedings of the 56th Annual Meeting of the Association
  for Computational Linguistics (Volume 1: Long Papers)}, pages 979--988,
  Melbourne, Australia. Association for Computational Linguistics.

\bibitem[{Yao et~al.(2019)Yao, Peng, Weischedel, Knight, Zhao, and
  Yan}]{yao2019plan}
Lili Yao, Nanyun Peng, Ralph Weischedel, Kevin Knight, Dongyan Zhao, and Rui
  Yan. 2019.
\newblock Plan-and-write: Towards better automatic storytelling.
\newblock In \emph{Proceedings of the AAAI Conference on Artificial
  Intelligence}, volume~33, pages 7378--7385.

\bibitem[{Yi et~al.(2020)Yi, Liu, Li, and Sun}]{ijcai2020-526}
Xiaoyuan Yi, Zhenghao Liu, Wenhao Li, and Maosong Sun. 2020.
\newblock \href {https://doi.org/10.24963/ijcai.2020/526} {Text style transfer
  via learning style instance supported latent space}.
\newblock In \emph{Proceedings of the Twenty-Ninth International Joint
  Conference on Artificial Intelligence, {IJCAI-20}}, pages 3801--3807.
  International Joint Conferences on Artificial Intelligence Organization.
\newblock Main track.

\bibitem[{Zhang* et~al.(2020)Zhang*, Kishore*, Wu*, Weinberger, and
  Artzi}]{bert-score}
Tianyi Zhang*, Varsha Kishore*, Felix Wu*, Kilian~Q. Weinberger, and Yoav
  Artzi. 2020.
\newblock \href {https://openreview.net/forum?id=SkeHuCVFDr} {Bertscore:
  Evaluating text generation with bert}.
\newblock In \emph{International Conference on Learning Representations}.

\bibitem[{Zhang et~al.(2019)Zhang, Wei, and Zhou}]{zhang-etal-2019-hibert}
Xingxing Zhang, Furu Wei, and Ming Zhou. 2019.
\newblock \href {https://doi.org/10.18653/v1/P19-1499} {{HIBERT}: Document
  level pre-training of hierarchical bidirectional transformers for document
  summarization}.
\newblock In \emph{Proceedings of the 57th Annual Meeting of the Association
  for Computational Linguistics}, pages 5059--5069, Florence, Italy.
  Association for Computational Linguistics.

\bibitem[{Zhu et~al.(2021)Zhu, Zhang, Liu, and Wang}]{zhu-etal-2021-neural}
Qingfu Zhu, Wei-Nan Zhang, Ting Liu, and William~Yang Wang. 2021.
\newblock \href {https://doi.org/10.18653/v1/2021.acl-long.339} {Neural
  stylistic response generation with disentangled latent variables}.
\newblock In \emph{Proceedings of the 59th Annual Meeting of the Association
  for Computational Linguistics and the 11th International Joint Conference on
  Natural Language Processing (Volume 1: Long Papers)}, pages 4391--4401,
  Online. Association for Computational Linguistics.

\end{thebibliography}
\bibliographystyle{acl_natbib}

\appendix

\section{Style-Specific Contents} \label{keywords}
We detail how we extract style-specific contents and explain how they are used from the following three aspects:

\paragraph{\textit{What do we mean by “style-specific content”?}}
 We refer to "style-specific content" as those mainly used in texts with specific styles and should be retained after style transfer. For example, "Harry Potter" and "Horcrux" are style-specific since they are used only in J.K. Rowling-style stories. When transferring J.K. Rowling-style stories to other styles, style-specific tokens shouldn't be changed. However, existing models tend to drop style-specific tokens since they are not trained to learn these tokens conditioned on other styles.
 
\paragraph{\textit{How do we extract “style-specific contents”?}}
We extract style-specific contents by (1) obtaining top-10 salient tokens using TF-IDF, (2) reserving only people names (e.g., "Harry Potter"), place names (e.g., "London"), and proper nouns (e.g., "Horcrux"), and (3) filtering out high-frequency tokens in all corpus (e.g., "London") since these tokens can be learned conditioned on every style. We regard the remaining tokens as style-specific contents. 

As mentioned before, we employ the TF-IDF algorithm on the corpus to obtain rough style-specific contents for different styles, respectively. The reason for using TF-IDF: it is necessary to ensure that the extracted tokens are salient to the story plots. We extract style-specific tokens from the salient tokens using the second and third steps. Then, we use a part-of-speech tagging toolkit (e.g., NLTK) to identify function words and prepositions to retain people's names, place names, and proper nouns. Note that the frequency is an empirical value observed from datasets. However, the TF-IDF algorithm chooses the important words corresponding to the special style based on word frequency. There may be some style-unrelated words that are important to the content. 
% We regard the words possessing the same frequency in different style corpus as style-unrelated words. 
Therefore, we need to filter out style-unrelated words. Concretely, we use Jieba\footnote{https://github.com/fxsjy/jieba}/NLTK\cite{bird2009natural} to collect the word frequency for Chinese and English datasets, respectively. Moreover, we regard the words possessing a high frequency in all styles corpus as style-unrelated words. Specifically, We set tokens appearing in 10\% samples in the dataset as high-frequency words. Then we filter out these words to obtain style-specific contents. The frequency value needs to be reset to apply the method to other datasets. 

\paragraph{\textit{How are the “style-specific contents” used?}}

One challenge of long-text style transfer is transferring discourse-level author style while preserving the main characters and storylines. It's difficult for existing models to transfer style-specific contents since they are not trained to learn these tokens conditioned on other styles. Therefore, we extract "style-specific contents" before style transferring and replace them with the special token "<Mask>". Then, the "style-specific contents" will be filled in the second stage, as shown in Figure~\ref{Fig_1}.

\begin{figure*}[t!]
\centering
\includegraphics[width=1\linewidth]{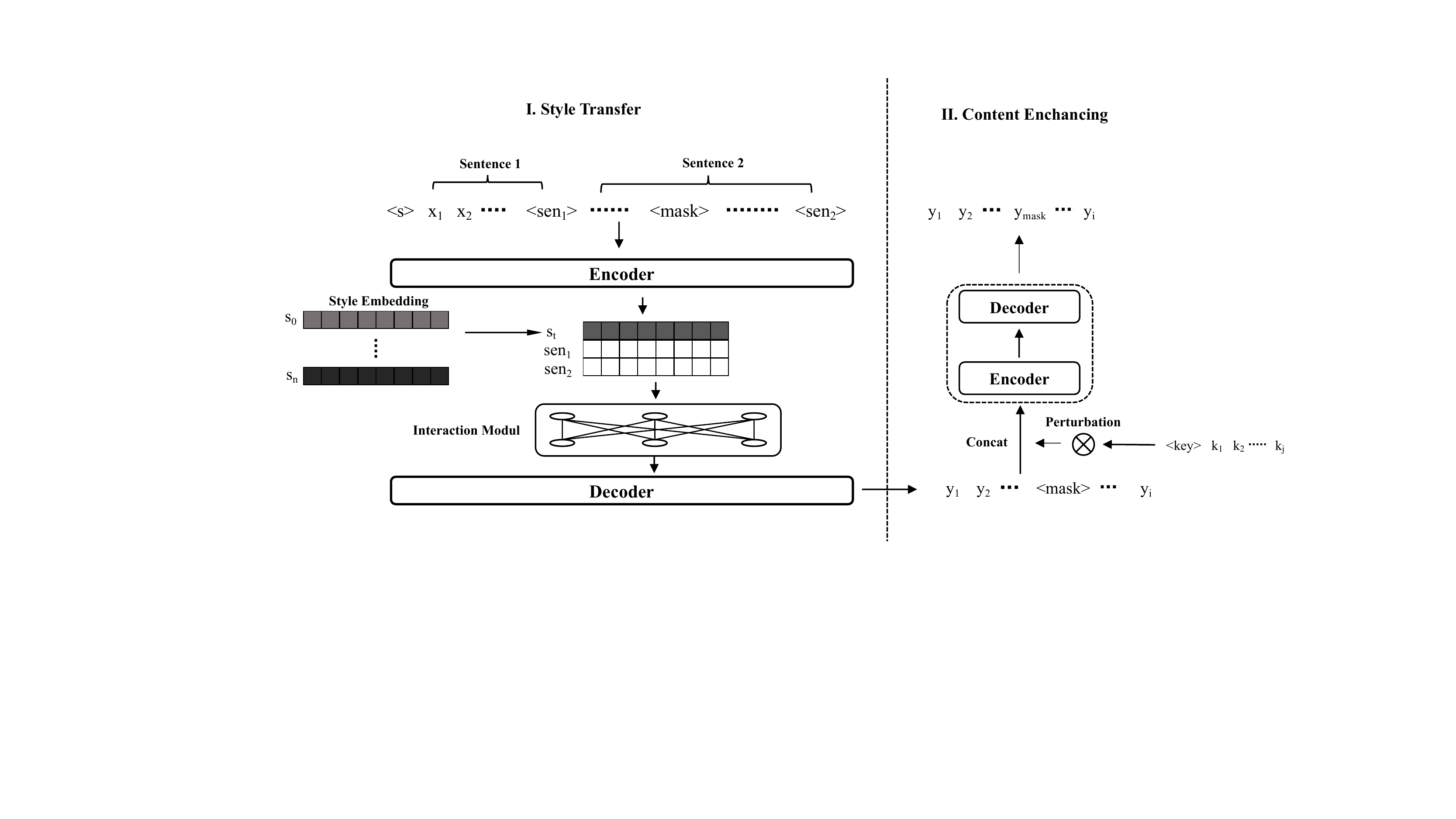}
\captionsetup{type=table}
\caption{More Chinese cases generated by baselines, which are transferred from the fairy tale style to the JY style. The number before the sentences indicate their corresponding sentences in the source text in semantics. The underline sentences indicate inserted content to align with target style. The English texts below the Chinese are translated versions of the Chinese samples.}
\label{more_case}
\end{figure*}

\section{Data Pre-Processing} \label{data_proprecessing}
Due to lack of stylized author datasets, we collected several authors' corpus to construct new datasets. As for Chinese, we extracted paragraphs from 21 novels of LuXun~(LX) and 15 novels of JinYong~(JY), and fairy tales collected by~\citet{guan2021lot}. On the other hand, the English dataset consists of everyday stories from ROCStories~\cite{mostafazadeh-etal-2016-corpus} and fragments from Shakespeare's plays. Each fragment of Shakespeare's plays comprises multiple consecutive sentences and as long as samples in ROCStories. We collect the Shakespeare-style texts from the Shakespeare corpus in Project Gutenberg\footnote{https://www.gutenberg.org} under the Project Gutenberg License\footnote{https://www.gutenberg.org/policy/license.html}. We use Jieba/NLTK~\cite{bird2009natural} for word tokenization for the Chinese/English dataset in data pro-processing. In addition, these data are public corpora, and we also check the information for anonymization.
% As illustrated in Figure~\ref{samples}, we split an entire novel into paragraphs, which contain consecutive sentences. 
% \begin{table}[!t]
%     \footnotesize
%     \centering
%     \scalebox{0.95}{
%     \begin{tabular}{cc|ccc|c|c}
    
\begin{figure}[ht]
\centering
\includegraphics[scale=0.35]{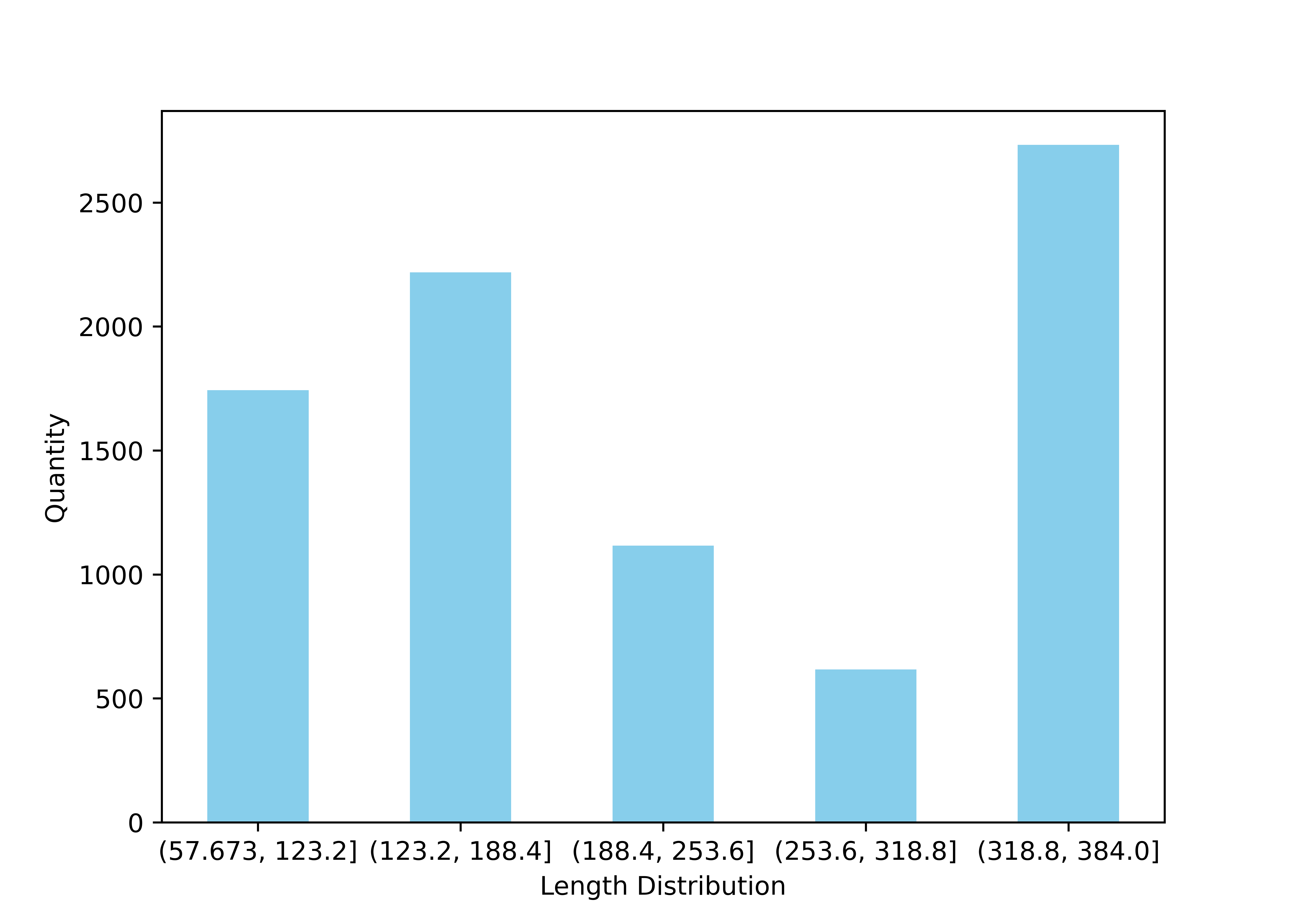}
\caption{Length distribution of texts in the Chinese dataset.}  
\label{Fig_data_sta}
\end{figure}

%     \toprule
%     \multicolumn{2}{c|}{\textbf{Dataset}} & \multicolumn{3}{c|}{\textbf{Train}} & \textbf{Val}& \textbf{Test}
%     \\
%     % &\multicolumn{3}{c}{\textbf{Chinese}} & \multicolumn{3}{c}{\textbf{English}} \\
%     % &\textbf{Train} & \textbf{Valid}  & \textbf{Test} &\textbf{Train} & \textbf{Valid}  & \textbf{Test} \\
%     \midrule
%     \multirow{3}{*}{\textbf{ZH}}
%     & \textbf{Style} & JY & LX &Tale & Tale & Tale\\
%     & \textbf{Size} & 2,964 & 3,036 & 1,456 & 242 & 729
%     \\
%     & \textbf{Avg Len} & 344 & 168 & 175 & 175 & 176 \\
%     \midrule
%     \multirow{2}{*}{\textbf{EN}}
%     & \textbf{Style} & \multicolumn{2}{c}{Shakespeare} &  ROC &ROC &ROC\\
%     & \textbf{Size} & \multicolumn{2}{c}{1,161}  & 1,161 & 290 &290 \\
%     & \textbf{Ave Len} & \multicolumn{2}{c}{71}  & 49 & 48 & 50\\
%   \bottomrule
%     \end{tabular}}
%     \caption{Statistics of the Chinese~(ZH) and English~(EN) datasets. Avg Len indicates the average length of tokens of each sample.} %the sizes and styles of train and test sets. 
%     %The tale in Chinese and story in English indicate fairy tales from LOT \cite{guan2021lot} and everyday stories from ROCStories \cite{mostafazadeh-etal-2016-corpus}, respectively.}
%     \label{data_sta}
% \end{table}
Regarding to limitation of modern language models, the length of samples is also limited. We set the max length as 384 and 90 for Chinese and English, respectively. Each sample has 4 sentences at least. We choose above length to balance the data length of different styles. Additionally, we filtered the texts which are too long to generate or too short to unveil author writing style. As Figure \ref{Fig_data_sta} shows, texts in the Chinese dataset spans a diverse range of length. 

\begin{table*}[!t]
\scriptsize
    \centering
    % \scalebox{0.73}{
    \begin{tabular}{lp{14cm}}
    \toprule
    \textbf{Authors} & \textbf{Texts}  \\
    \midrule
    \textbf{JY} 
    & 
    \begin{CJK*}{UTF8}{gkai}
    杨过左手抢过马缰，双腿一夹，小红马向前急冲，绝尘而去。郭芙只吓得手足酸软，慢慢走到墙角拾起长剑，剑身在墙角上猛力碰撞，竟已弯得便如一把曲尺。以柔物施展刚劲，原是古墓派武功的精要所在，李莫愁便拂尘、小龙女使绸带，皆是这门功夫。杨过此时内劲既强，袖子一拂，实不下于钢鞭巨杵之撞击。杨过抱了郭襄，骑着汗血宝马向北疾驰，不多时便已掠过襄阳，奔行了数十里，因此黄蓉虽攀上树顶极目远眺，却瞧不见他的踪影。
    \end{CJK*}
    
    Yang Guo grabbed the horse's reins with his left hand, clamping with his leg, and then little red horse rushed out of sight. Guo Fu was so frightened that his hands and feet were sore, and she slowly walked to the corner to pick up the long sword. Using soft objects to display strength was originally the essence of the ancient tomb school martial arts. Yang Guo's internal energy was strong at this moment, and a flick of his sleeve was no less than the impact of a giant steel whip. Yang Guo hugged Guo Xiang, and rode a sweaty horse to the north. After a while, he passed Xiangyang and ran for dozens of miles. Although Huang Rong climbed to the top of the tree and looked far into the distance, she could not see any trace of him.
    \\
    \midrule
    \textbf{LX}
    &
    \begin{CJK*}{UTF8}{gkai}
    自《新青年》出版以来，一切应之而嘲骂改革，后来又赞成改革，后来又嘲骂改革者，现在拟态的制服早已破碎，显出自身的本相来了，真所谓“事实胜于雄辩”，又何待于纸笔喉舌的批评。所以我的应时的浅薄的文字，也应该置之不顾，一任其消灭的；但几个朋友却以为现状和那时并没有大两样，也还可以存留，给我编辑起来了。这正是我所悲哀的。我以为凡对于时弊的攻击，文字须与时弊同时灭亡，因为这正如白血轮之酿成疮疖一般，倘非自身也被排除，则当它的生命的存留中，也即证明着病菌尚在。
    
    Since the publication of "New Youth", everyone has ridiculed the reform in response to it, later approved of it, and then ridiculed the reformers. Now the mimetic uniform has long been broken, showing its true nature. The so-called "facts speak louder than words", why should they be criticized by pen and paper mouthpieces. Therefore, my timely and superficial writing should also be ignored and wiped out. However, a few friends thought that the current situation was not much different from that at that time, and they could still be preserved, so they edited them for me. This is what I am saddened by. I think any attack on the evils of the times, the writing must perish at the same time as the evils of the times, because this is like the boils and boils caused by the white blood wheel. If it is not eliminated by itself, the existence of its life also proves that the germs are still there.
    \end{CJK*} 
    \\
    \midrule
    \textbf{Tale}
    &
    \begin{CJK*}{UTF8}{gkai}
    有个财主，非常喜欢自家的一棵橘子树。谁从树上摘下一个橘子，他就会诅咒人家下十八层地狱。这年，橘子又挂满了枝头。财主的女儿馋的直流口水。忍不住摘了一个，刚尝了一口，就不省人事了。财主后悔不已，把树上的橘子都摘下来，分给邻居和路人。最后一个橘子分完，女儿就苏醒了。财主再也不敢随便诅咒别人了。
    
    There was a rich man who liked his orange tree very much. Whoever plucks an orange from the tree, he will curse him to eighteen levels of hell. This year, oranges are hanging on the branches again. The rich man's daughter was drooling. Then, she couldn't help picking one, and just after a bite, she was unconscious. The rich man was remorseful, so he plucked all the oranges from the tree and gave them to neighbors and passers-by. After the last orange was given, the daughter woke up. The rich man no longer dared to curse others casually.
    \end{CJK*} 
    \\
    \midrule
    \textbf{ROC}
    &
    Garth has a chicken farm. Each morning he must wake up and gather eggs. Yesterday morning there were 33 eggs! After gathering the eggs, he feeds the chickens. Finally he gets to eat breakfast, and go to school.
    \\
    \midrule
    \textbf{Shakespeare}
    &
    King. Giue them the Foyles yong Osricke, Cousen Hamlet, you know the wager. Ham. Verie well my Lord, Your Grace hath laide the oddes a 'th' weaker side. King. I do not feare it, I haue seene you both: But since he is better'd, we haue therefore oddes. Laer. This is too heauy, Let me see another.
    \\
    \bottomrule
    \end{tabular}
    % }
    \caption{Samples of different authors in Chinese and English datasets. The English texts below the Chinese are translated versions of the Chinese samples.}
    \label{samples}
\end{table*}

% \section{Case Study}
% We perform case studies for better understanding the model performance. Table~\ref{case_study} shows generated texts from different models. Obviously, StyleLM can only rewrite individual sentences and simply merge some content. Although it can also expand some content. On the contrary, our LongTrans model can not only rewrite most sentences but also supplement the plots to be more in line with the target style.

\section{Different Style Samples}\label{sample_analyse}
In process of constructing datasets, we try to collect different author corpus who have a gap in writing styles. As shown in Table~\ref{samples},  the JY-style texts mostly describe martial arts actions and construct interesting plots, while the LX-style texts focus on realism with profound descriptive and critical significance. And the fairy tales differ from these texts in terms of topical and discourse features. In the English datasets, the Shakespeare-style texts are flamboyant and contain elaborate metaphors and ingenious ideas, which the everyday stories are written in plain language and without rhetoric.

% \begin{figure}[t!]
% \centering
% \includegraphics[width=0.9\linewidth]{./figure/style_golden_zh.pdf}
% \caption{Visualization of the golden Chinese test dataset on style space using t-SNE. } 
% \label{style_golden_zh}
% \end{figure}

\section{More Implementation Details}\label{more_implementation_details}
In terms of selecting pre-trained model, LongLM$_{\rm base}$ and T5$_{\rm base}$ are the generic base model for the Chinese and English generation, respectively. To optimize the models for these specific languages, we have fine-tuned them using different hyperparameter values ($\lambda_{1/2/3}$). These values were determined based on the performance observed on a validation set, which was created by pre-extracting 5\% of the training data for this purpose.

% Different hyper-parameter values ($\lambda_{1/2/3}$) for English and Chinese datasets: We chose the corresponding hyperparameter values based on the performance of the validation set. And 5\% of traninig set has been pre-extracted as the validation set. 

\section{More Ablation Study Results} \label{more_ablation}
To explore the effect of the proposed component, we also conduct more ablation studies on Chinese datasets. As shown in Table~\ref{zh_ablation}, the ablation of $\mathcal{L}_{\rm dis}$ leads to better style accuracy, which show the different trends comparing with English dataset. We conjecture that $\mathcal{L}_{\rm dis}$ aims to maintain the content and reduces style information. Without $\mathcal{L}_{\rm dis}$, the powerful $\mathcal{L}_{\rm style}$ leads the StoryTrans to degenerate to style conditional language model. Furthermore, the ablation of $\mathcal{L}_{\rm style}$ also confirms the powerful ability of style control as in previous paper. And we find that when removing $\mathcal{L}_{\rm sop}$, the model loses the ability to transfer at the discourse level and has only learned token-level copy.

\begin{table*}[!t]
    \scriptsize
    \centering
    \begin{tabular}{c|l|ccccccc|cc}
    \toprule
    \textbf{Target Styles}
    &\textbf{Model}
    & \textbf{r-Acc}&\textbf{a-Acc} & \textbf{BLEU-1} & \textbf{BLEU-2} & \textbf{BS-P} & \textbf{BS-R} & \textbf{BS-F1} & \textbf{BL-Overall} & \textbf{BS-Overall}

    \\
    \midrule

\multirow{4}{*}{\textbf{ZH-LX}}
&\textbf{Proposed Model} &\text{97.66} &\text{59.94} & \text{32.19} & \text{14.44} & \text{68.53} & \text{70.48} & \text{69.45} & \text{37.38} & \text{64.52} \\

\cline{2-11}

&\textbf{(-)~$\mathcal{L}_{\rm dis}$}& \text{99.86} &\text{92.59} & \text{20.36} & \text{5.45} & \text{63.37} & \text{62.96} & \text{63.14} & \text{34.56} & \text{76.46} \\

&\textbf{(-)~$\mathcal{L}_{\rm style}$}&\text{88.06} & \text{12.20} & \text{43.09} & \text{23.88} & \text{75.44} & \text{75.68} & \text{75.53} & \text{20.21} & \text{30.35} \\

&\textbf{(-)~$\mathcal{L}_{\rm sop}$} &\text{87.10} &\text{2.05} & \text{54.38} & \text{32.95} & \text{81.19} & \text{79.77} & \text{80.42} & \text{9.46} & \text{12.83} \\

\midrule

\multirow{4}{*}{\textbf{ZH-JY}}
&\textbf{Proposed Model}&\text{84.49} & \text{62.96} & \text{30.71} & \text{14.5} & \text{68.76} & \text{71.69} & \text{70.16} & \text{37.72} & \text{66.46} \\

\cline{2-11}

&\textbf{(-)~$\mathcal{L}_{\rm dis}$}& \text{97.53} &\text{92.59} & \text{18.49} & \text{4.85} & \text{62.17} & \text{65.42} & \text{63.73} & \text{32.87} & \text{76.81} \\

&\textbf{(-)~$\mathcal{L}_{\rm style}$}&\text{61.86} & \text{40.87} & \text{39.78} & \text{21.97} & \text{73.73} & \text{75.42} & \text{74.52} & \text{35.52} & \text{55.18} \\

&\textbf{(-)~$\mathcal{L}_{\rm sop}$} &\text{61.72} &\text{10.83} & \text{51.29} & \text{30.98} & \text{79.65} & \text{79.82} & \text{79.72} & \text{21.10} & \text{29.38} \\

% \textbf{(-)~CE} &\text{92.41} &\text{73.10} & \text{21.62} & \text{6.09} & \text{79.73} & \text{86.12} & \text{82.59} & \text{31.82} & \text{77.70} \\

  \bottomrule 
    \end{tabular}
    \caption{More ablation study results on Chinese datasets.~(-) indicates removing the component in proposed model. ZH-LX/ZH-JY is the Chinese author LuXun/JinYong, respectively. 
    }
    \label{zh_ablation}
\end{table*}

\section{Style Analysis of Transferred Texts} \label{apd:style_analaysis}

\begin{figure}[t!]
\centering
\includegraphics[width=0.8\linewidth]{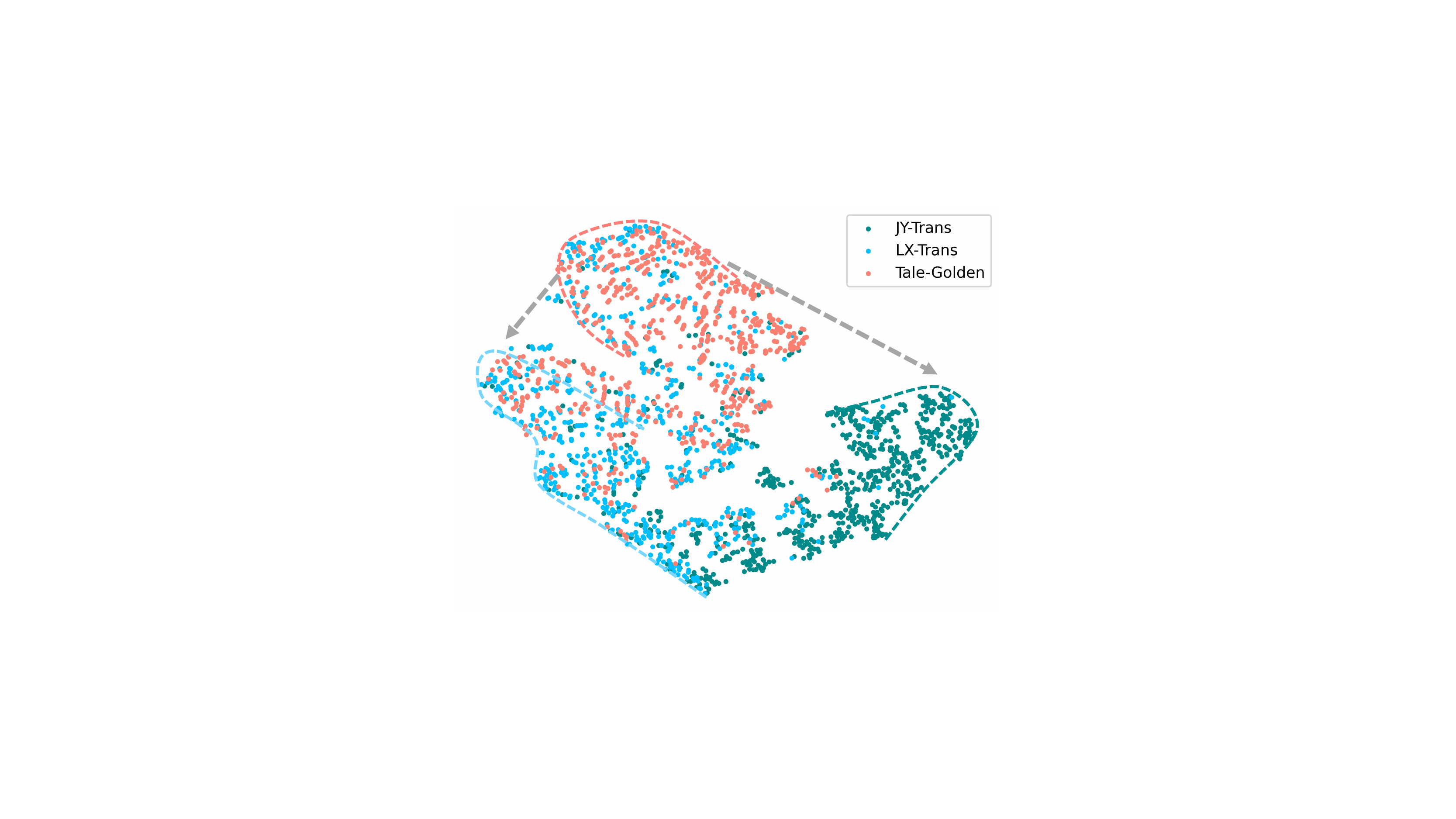}
\caption{Visualization of the golden LX-style texts and transferred LX-style texts on style space using t-SNE. } 
\label{lx_vis}
\end{figure}

In order to investigate whether our StoryTrans indeed rephrase the expression of texts, we employ surface elements of text to show author writing styles. And the surface element are associated with statistical observations. For example, the small average length of sentences show the author preference to write a short sentence, and more question marks indicate the author accustomed to using questions. To this end, we use the number of (1) commas, (2) colons, (3) sentences in a paragraph, (4) question mark (5) left quotation mark, (6) right quotation mark, and (7) average number of words in a sentence to quantify surface elements into a 7 dimension vector. Then we leverage the t-SNE to visualize the golden texts and transferred texts. As shown in Figure~\ref{style_analysis}, different style distribute separately across the style space. This proves JY, LX and fairy tale in Chinese dataset have a gap in writing style. And Figure~\ref{lx_vis} shows the transferred texts fall in golden texts in style space, indicating StoryTrans successfully transferred the writing style.
 
\section{More Details of Manual Evaluation} \label{ana_manual_eva}
In addition to automatic evaluation, we conduct manual evaluation on generated texts. 
% We define coherence as the intra-sentence fluency. 
As mentioned before, we require the annotators to score each aspect from 1~(the worst) to 3~(the best). As for payment, we pay 1.8 yuan (RMB) per sample in compliance with Chinese wage standards. Our annotators consist of undergraduate students who are experienced in reading texts written in the styles of the respective authors (JY and LX). To ensure they fully understand the evaluation metrics, we conducted case analyses with them. Our scoring rubric assigns 1, 2, or 3 points to the transferred text based on the proportion of sentences meeting the following criteria (1/3, 2/3, or 3/3):
\begin{itemize}
    \item Style Accuracy: whether the transferred text conforms to the corresponding style.
    \item Content Preservation: whether the source content, such as character names, are retained.
    \item Coherence: whether the sentences in the transferred text are semantically connected.
\end{itemize}
And we compute the final score of each text by averaging the scores of three annotators.

As illustrated in manual evaluation,
we observe that the results mainly conform with the automatic evaluation. Our StoryTrans obtained the highest score on the style accuracy in both transferred directions by a sign test compared to the other baselines, showing its stable ability of style control. Moreover, in terms of content preservation, the score of StoryTrans is comparable with StyleLM and slightly higher than Reverse Attention, demonstrating that StoryTrans can keep the main semantics of input. In terms of coherence, the score of StoryTrans is also comparable with baselines, showing some room for improvement. As discussed before, Style Transformer tends to copy the input, leading to the highest performance in content preservation and coherence. In summary, human evaluation depicts the strength of StoryTrans not only on style control but also on overall performance, indicating a balance of these metrics.

\section{More Case Studies}\label{more_case_study}
We show more cases in Table~\ref{more_case}. Comparing source text with Style Transformer, Style Transformer copies the input and only changes little tokens. This result also confirms with highest BLEU and BERTScore in automatic results. Like StyleLM, Reverse Attention also incorporates some target author content into generated texts. However, Reverse Attention inserts too much content that overwhelms original plots. Furthermore, some critical entities~(e.g., character name, \begin{CJK*}{UTF8}{gkai}{“柯里教授”}\end{CJK*}~/“Professor Curry”~$\rightarrow$~\begin{CJK*}{UTF8}{gkai}{“柯镇恶”}\end{CJK*}~/“Ke Zhen'e”) are revised to the similar word on in target author corpus.
% , which reveal effective transfer ability in token level.
To maintain the story coherence, these important entities should stay the same. 
In summary, the token-level transfer may destroy the essential plots and damage the coherence.

\end{document}